\documentclass[10pt,journal,compsoc]{IEEEtran}

\usepackage{color}
\usepackage{times}
\usepackage{epsfig}
\usepackage{graphicx}
\usepackage{amsmath}
\usepackage{amssymb}
\usepackage{multirow}
\usepackage{floatrow}
\usepackage{mathrsfs}
\usepackage{array}
\usepackage{ifpdf}
\usepackage{algorithmic}
\usepackage{algorithm}
\usepackage{url}

\ifCLASSINFOpdf
  \DeclareGraphicsExtensions{.pdf,.jpeg,.png}
    \DeclareGraphicsExtensions{.eps}
\usepackage{epstopdf}
\else
\fi

\newcommand{\yukai}[1]
{\textcolor{black}{#1}}

\newcommand{\syk}[1]
{\textcolor{black}{#1}}

\newfloatcommand{capbtabbox}{table}[][\FBwidth]

\begin{document}
%\title{Structured Scene Parsing by Learning CNN-RsNN with Sentence Descriptions}
\title{Face Hallucination by Attentive Sequence Optimization with Reinforcement Learning}


\author{Yukai Shi, Guanbin Li, Qingxing Cao, Keze Wang, Liang Lin

\thanks{This work was supported in part by the National Key Research and Development Program of China under Grant
No.2018YFC0830103 and No.2016YFB1001004, in part by the NSFC-Shenzhen Robotics Projects (U1613211), in part by the National Natural Science Foundation of China under Grant No.61702565, in part by National High Level Talents Special Support Plan (Ten Thousand Talents Program) and in part by the Fundamental Research Funds for the Central Universities under Grant No.18lgpy63. \emph{(Corresponding author: Guanbin Li).}}
\thanks{Y. Shi, G. Li, Q. Cao, K, Wang and L. Lin are with the School of Data and Computer Science, Sun Yat-Sen University, 510006, China (e-mail: \{shiyk3,caoqx\}@mail2.sysu.edu.cn; \{liguanbin\}@mail.sysu.edu.cn; kezewang@gmail.com; linliang@ieee.org).}
}

\markboth{IEEE TRANSACTIONS ON PATTERN ANALYSIS AND MACHINE INTELLIGENCE}%
{Shell \MakeLowercase{\textit{et al.}}: Bare Demo of IEEEtran.cls for IEEE Transactions on Magnetics Journals}

\IEEEcompsoctitleabstractindextext{
\begin{abstract}
Face hallucination is a domain-specific super-resolution problem that aims to generate a high-resolution~(HR) face image from a low-resolution~(LR) input. In contrast to the existing patch-wise super-resolution models that divide a face image into regular patches and independently apply LR to HR mapping to each patch, we implement deep reinforcement learning and develop a novel attention-aware face hallucination (Attention-FH) framework, which recurrently learns to attend a sequence of patches and performs facial part enhancement by fully exploiting the global interdependency of the image. Specifically, our proposed framework incorporates two components: a recurrent policy network for dynamically specifying a new attended region at each time step based on the status of the super-resolved image and the past attended region sequence, and a local enhancement network for selected patch hallucination and global state updating. The Attention-FH model jointly learns the recurrent policy network and local enhancement network through maximizing a long-term reward that reflects the hallucination result with respect to the whole HR image. Extensive experiments demonstrate that our Attention-FH significantly outperforms the state-of-the-art methods on in-the-wild face images with large pose and illumination variations.
%Therefore, our proposed Attention-FH is capable of adaptively inferring an optimal searching path for each face image according to its own appearance feature.
\end{abstract}

}

\maketitle

\IEEEdisplaynontitleabstractindextext

% \fi
\IEEEpeerreviewmaketitle

\section{Introduction}
\label{intro}

Face hallucination, a.k.a. facial image super-resolution, aims to generate a high-resolution~(HR) face image from a given low-resolution~(LR) input. Face hallucination is a fundamental problem in the  field of face analysis and has drawn considerable research attention due to the need for solving this problem in various face-related tasks, including face attribute recognition~\cite{liu2015deep}, face alignment~\cite{zhang2016learning,liu2019facial} and face recognition~\cite{zhou2015naive}, under complex low-quality real-world scenarios. 

As a special form of general image super-resolution, obvious structural prior information exists in face images, which is therefore widely used in existing face hallucination algorithms~\cite{ledig2016photo,chen2017fsrnet,zhu2016deep}. Face prior information is usually embedded into the existing face hallucination models in the form of face component analysis~\cite{song2017learning}, facial correspondence field~\cite{zhu2016deep} and facial landmark localization~\cite{chen2017fsrnet}. However, the calculation of this prior information requires additional calculations, and accurate parsing of the landmarks is difficult in low-resolution situations.
%SENIOR EDITOR: Please ensure that the intended meaning has been maintained in this edit.
 Therefore, there is a series of work attempts to replace the fine-grained face prior computation with rough patch-wise super-resolving mapping, which can improve the efficiency of the algorithm while achieving comparable performance. Due to differences in appearance of facial organs and the natural symmetry of the facial region, existing patch-wise face hallucination methods either extract patches from detected facial landmarks or simply divide the face image into even patches and then independently perform LR to HR mapping on each detected patch~\cite{yang2010image,yang2013structured,ma2010hallucinating,tappen2012bayesian}. Specifically, end-to-end deep convolutional networks~(CNNs) have recently achieved great success in learning discriminative patch-to-patch mapping from LR images to HR images~\cite{bicnn,zhu2016deep}. However, face structure priors and spatial configurations~\cite{Liu2007,dong2014learning} are often treated as external information, and the contextual dependencies among the super-resolution reconstruction of each patch are usually ignored during the hallucination processing.

The difficulty of super-resolution reconstruction also varies due to the inconsistency in the degree of detail deterioration of each facial part. On the other hand, the symmetry of the human face and the similarity in the appearance of the adjacent regions make previously hallucinated HR patches worthy of reference to the latter. Therefore, during the process of face hallucination, the reconstruction sequence of patches and the selection of patch locations at each step are crucial for global face hallucination, which is consistent with the human visual perception mechanism. When people observe a scene object, they usually start with perceiving the whole image and successively explore a sequence of regions via the attention shifting mechanism rather than separately processing the local regions. This finding motivates us to explore a new pipeline for face hallucination by sequentially searching for the local attention regions and considering their contextual dependency from a global perspective. 

Inspired by the effectiveness of recurrent visual attention modeling for visual analysis and understanding~\cite{sun2015deepid3,wang2017multi,chen2018recurrent,li2019cross}, we propose an attention-aware face hallucination (Attention-FH) framework that fully exploits the global contextual information of the face to recurrently discover and enhance a series of local face regions. Specifically, accounting for the diverse characteristics of face images in terms of blur, pose, illumination and facial appearance, we model the face hallucination problem as a strategy optimization problem for patch sequence selection and implement a search for an optimal enhancement route. We resort to the deep reinforcement learning (RL) model~\cite{alphaGo} to sequentially determine local patches for enhancement, as the RL technique has shown promising results on decision-making problems without the need for supervision information at each step.

Specifically, our Attention-FH framework jointly optimizes
a recurrent policy network that learns to identify the facial region to be hallucinated at each time step and a local enhancement network for facial part super-resolution by considering the whole face image with previously enhanced parts. In our framework, rich correlation cues among different facial parts are explicitly exploited to guide the current region assignment, while past hallucination results are incorporated as a global reference during local enhancement in each step. In this way, the agent can make full use of the symmetry of the human face and the adjacent regions to assist in obtaining more accurate facial part hallucination reconstruction. For example, the agent can improve the enhancement of the right eye region by taking a clear version of the left eye region as reference. 

Instead of performing supervision in each step, we employ a single global reward for RL, which measures the overall performance of the entire hallucinated HR face. The optimization of the recurrent policy network is updated following the RL algorithm~\cite{REINFORCE}, which can be treated as a Markov decision process (MDP) maximized with a long-term global reward. At each time step, the policy network learns to determine the location and the size of an optimal rectangular facial region by conditioning on the whole face image with all previously enhanced results and the encoded action history. A gated recurrent unit (GRU) layer is employed to encode the information from the previously attended facial regions. All the previously determined regions are also recorded to avoid duplicated selection of a region in a recurrent mode. 

Given the attended facial region in each step, the local enhancement network is trained for hallucination reconstruction, and its loss is defined as the $L_2$ distance between the part hallucination result and the specific ground truth. Compared with whole-face super-resolution, the structure of facial components~(attended parts) is stable and easy to restored. Notably, the supervision information from the enhancement of facial parts also effectively reduces unnecessary trial and error during the reinforcement optimization.

We compare the proposed Attention-FH approach with state-of-the-art face hallucination methods under both constrained and unconstrained settings. Extensive experiments show that our method substantially outperforms all the alternative methods. Moreover, our framework can explicitly generate a sequence of attentional regions during the hallucination, which finely accords with the human perception process and to some extent provides an interpretable mechanism in the hallucination recovery process.

A preliminary version of this work is published in~\cite{cao2017attention}.  In this work, we inherit the idea of exploring the interdependency of facial components and redevelop the policy network from the perspective of the attention mechanism and reinforcement optimization. The improvement upon the initial version includes size-free attention and a new reward function designed to maximize the stability of the RL. We have also added a comprehensive discussion of the design of the local enhancement network and greatly improved its efficiency while maintaining its performance. Moreover, we present more comparisons with state-of-the-art models and a more comprehensive ablation study on our proposed framework. 

The rest of this paper is organized as follows. Section~\ref{sec:relatedwork} reviews related work on face hallucination and deep RL. In Section~\ref{sec:method},
we introduce our proposed Attention-FH model. Section~\ref{sec:exper} provides extensive performance evaluation and comparisons with state-of-the-art models. Finally, we
conclude this paper in Section~\ref{sec:conclude}. 

\begin{figure*}[t]
    \centering
    \includegraphics[width=1\textwidth]{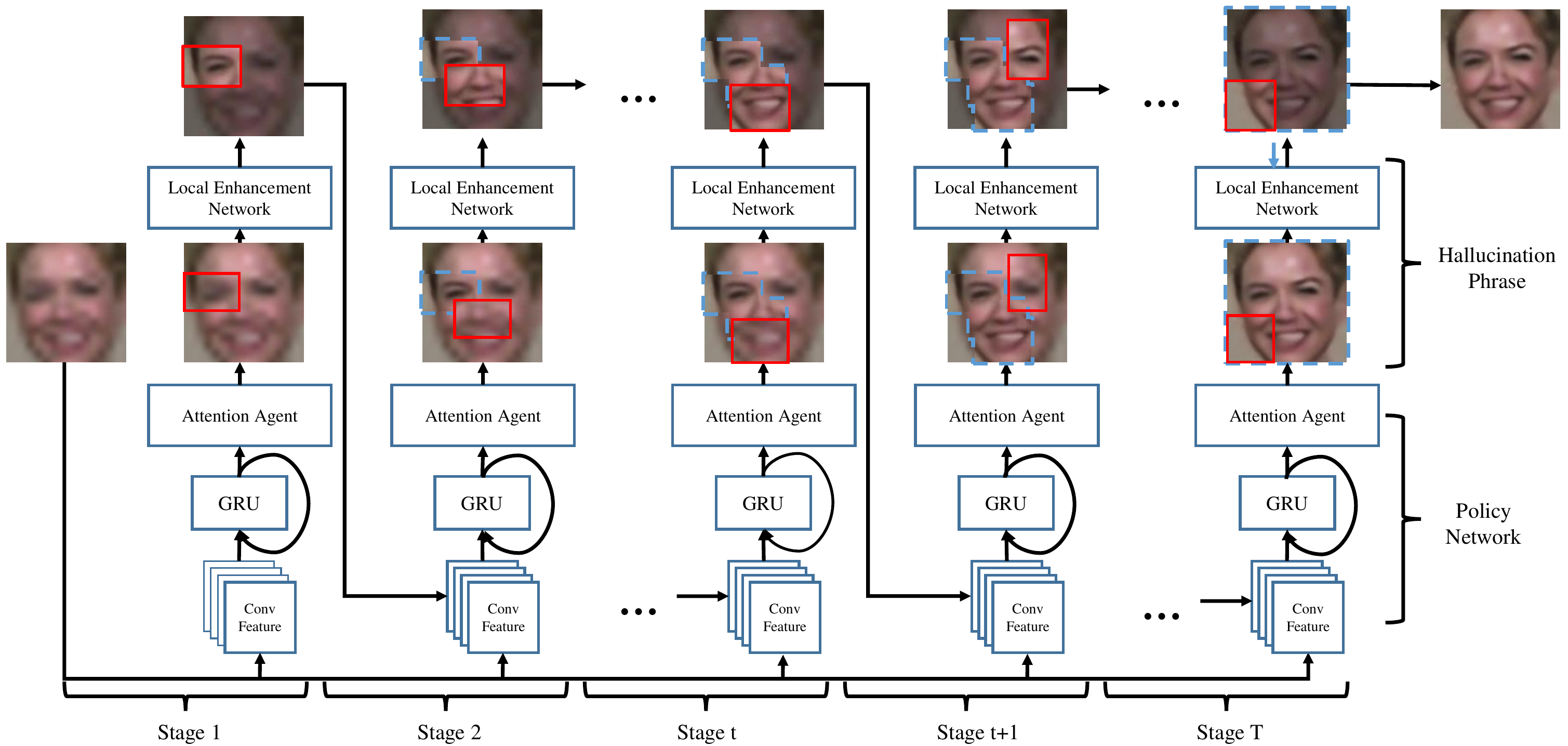}
    \caption{Pipeline of Attention-FH. The proposed framework contains two modules: the policy network and the local enhancement network. In each step, the attention agent glimpses the whole image and provides actions. Each action indicates the center position of the next attended rectangular region and the size of the bounding box. The attended patch is further fed as input to a local enhancement network for super-resolution. We use the red solid bounding boxes and the blue dashed bounding boxes to indicate the attended patch and enhanced patch, respectively. A global reward is given at the end of the sequence to enforce the policy network to learn the optimal restoring route. With the two components, we can perform a coarse-to-fine face hallucination paradigm.}
    \label{fig:intro}
    \vspace{2mm}
\end{figure*}

\section{Related Work}\label{sec:relatedwork}

\textbf{Face Hallucination and Image Super-Resolution} Face hallucination is a domain-specific image super-resolution problem proposed to map a LR facial image to its HR version. With obvious prior knowledge, face hallucination methods are required to handle extremely degraded faces and restore complex structural information. Early approaches hypothesized that corrupted faces are in a relatively controlled environment with small variations. Yang \emph{et~al.}~\cite{yang2010image} enforced LR and HR images to have similar sparse representations and implemented image super-resolution by taking into account the sparsity prior. {\yukai{Wang \emph{et~al.}~\cite{wang2005hallucinating} decomposed faces into different frequency bands and hallucinated faces by eigentransformation. By contrast, structured 
%SENIOR EDITOR: Please consider whether ``structured'' rather than ``structure'' was intended here.
face hallucination (SFH)~\cite{yang2013structured} pre-aligns faces and establishes the mapping between facial components.} SFH not only achieves impressive results but also reveals that facial components are crucial in face hallucination. However, SFH relies heavily on pre-alignment, making it difficult to cope with situations where illumination and pose changes are considerable. More recently, deep neural networks have shown impressive performance in face hallucination and image restoration~\cite{vdsr,gln,dong2014learning,zeroshot,bicnn}. Dong \emph{et~al.}~\cite{dong2014learning} employed a fully convolutional network (FCN) for image SR. Kim \emph{et~al.}~\cite{vdsr} made a very deep CNN for image SR trainable by adopting a highway connection. \syk{Reed \emph{et~al.}~\cite{reed2017parallel} proposed an efficient sampling strategy to demonstrate high-quality image reconstruction. Ulyanov~\emph{et~al.}~\cite{imageprior} further demonstrated that deep neural networks can make full use of prior knowledge of the image itself to rebuild a corrupted image.} Shocher~\emph{et~.al.}~\cite{zeroshot} employed a novel internal learning approach to fully explore LR inputs. UR-DGN~\cite{yu2016ultra} is claimed to be the first face SR method that uses a generative adversarial network. Tuzel~\emph{et~.al}~\cite{gln} claimed that global information is crucial to face hallucination and established a local and global network to restore faces.

However, all these existing deep learning models attempt to improve the performance of face hallucination by designing deeper and more complex neural network structures. In this work, we start from the perspective of human cognition and model the face hallucination process as a patch-wise local reconstruction problem. We introduce a deep RL-based optimization method to learn a series of ordered patch hallucination sequences. For patch-wise hallucination,which is not our main focus, we draw on an existing FCN-based framework and incorporate it into our Attention-FH model for end-to-end training. We firmly believe that the Attention-FH framework proposed in this paper is compatible with any existing deep super-resolution models and will benefit from the future improvement of super-resolution algorithms.

%\subsection{Recurrent generative model}
%The recurrent model has been recently proved to be effective
%\cite{recurrentSR} \cite{recurrentSaliency}

\textbf{Attention and Reinforcement Learning} Visual attention modeling is inspired by the human visual perception system. Visual attention modeling is widely embedded in existing deep neural networks in the form of adaptive feature weighting or salient region localization and has been proved to be effective in improving the performance of a series of computer vision tasks, including object proposal\cite{NIPS2016_6532}, object classification\cite{NIPS_attention}, relationship detection\cite{relDet}, image captioning\cite{captionAttn} and visual question answering\cite{vqa}. Some works have exploited RL to optimize the attention networks to address the problem that the coordinates of attended regions are not differentiable.
For example, \cite{FaceDetRL} and \cite{caicedo2015active} learned an agent that actively locates the target regions (face or objects) instead of exhaustively sliding subwindows on images. Goodrich \emph{et~al.}~\cite{FaceDetRL} defined 32 actions to shift the focal point and reward the agent when spotting the target. Caicedo \emph{et~al.}~\cite{caicedo2015active} defined an action set that contains several transformations of the bounding box and rewarded the agent if the bounding box became closer to the ground truth in each step. Both of these methods learned an optimal policy to locate the target through Q-learning.

\section{Methodology}\label{sec:method}
\begin{figure*}[t]
    \centering
    \includegraphics[width=1\textwidth]{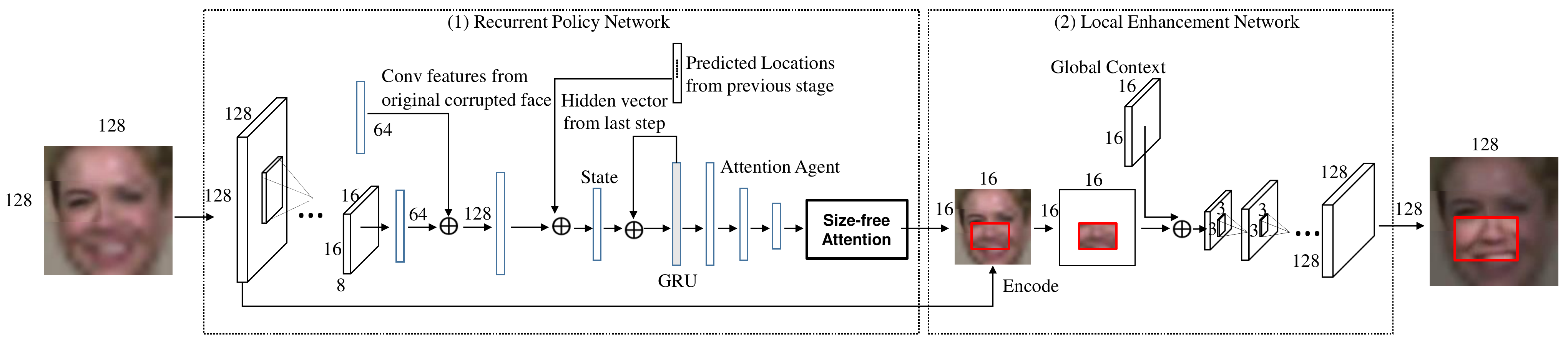}
    \caption{Network architecture of our Attention-FH. At each time step $t$, the recurrent policy network outputs the actions by observing the original corrupted face image $I_0$, the current enhanced face image $I_{t-1}$ and the historical actions $v_l$. Meanwhile, $v_l$ is encoded by latent variables (e.g.~64 hidden states) of GRU. $I_0$ and $I_{t-1}$ are first represented as high-level features from an output feature vector of a fully connected layer and are then concatenated to form a vector of 128 dimensions. The above three corresponding pieces of information constitute the state $s$. The GRU layer learns to infer the action probabilities by considering $s$. The output probabilities are formulated by a fully connected linear layer~(128 neurons) for all candidate actions. Given the actions, we can obtain the local patch $\hat{I}_{t-1}^{l_t}$. Then, $\hat{I}_{t-1}^{l_t}$ is sent to the local enhancement network for hallucinating, which results in the enhanced patch $\hat{I}_{t}^{l_t}$. Finally, a new hallucinated face is generated by replacing $\hat{I}_{t-1}^{l_t}$ with $\hat{I}_{t}^{l_t}$.% Network architecture of our recurrent policy network and local enhancement network. At each time step, the recurrent policy network takes \yukai{the} current hallucination result $I_{t-1}$ and \yukai{the} action history vector encoded by \qingxing{GRU} (128 hidden states) as the input and then outputs the action probabilities for all $W\times H$ locations, where W, H are the height and width of the input image. The policy network first encodes $I_{t-1}$ with one fully-connected layer (256 neurons), and then fuse the encoded image and the action vector with \qingxing{an GRU layer}. Finally, a fully-connected linear layer is appended to generate the $W\times H$-way probabilities. Given the probability map, we extract the local patch, then pass the patch and $I_{t-1}$ into the local enhancement network to generate the enhanced patch. The local enhancement network is constructed by two fully-connected layers (each with 256 neurons) for encoding $I_{t-1}$ and $8$ cascaded convolutional layers for image patch enhancement. Thus a new face hallucination result can be generated by replacing the local patch with an enhanced patch.
    }
    \label{fig:model_overview}
    \vspace{2mm}
\end{figure*}
\subsection{Inference Overview}
We develop an Attention-FH to perform face hallucination in a coarse-to-fine manner. Specifically, Attention-FH is composed of two parts: a recurrent policy network that learns to adaptively locate a particular facial region for local hallucination, and a local enhancement network that directly learns to map a located facial patch with LR to its HR version, considering both the local and global perspective.

Given a face image $I_{lr}$ with LR, the target of our proposed Attention-FH framework is to generate the corresponding HR version $I_{hr}$ through a series of iterative local patch enhancements, which can be formulated as:
\begin{equation}
I_{hr} = F(I_{lr}|\theta),
\end{equation}
where $\theta$ denotes the parameters of the Attention-FH model, and $F$ is the whole hallucinating procedure.

Given a state $s$, the recurrent policy network is learned to predict the actions $l_t$, including the center position of the next attended rectangular region and the size of the bounding box. The attended patch is further cropped and fed as input to a local enhancement network for super-resolution. This process can be formulated as:
\begin{equation}
\begin{aligned}
l_t = f_\pi(s_{t-1};\theta_\pi), \\
\hat{I}_{t-1}^{l_t} = g(l_t,I_{t-1}),
\end{aligned}
\end{equation}
where $\theta_\pi$ is the parameters of the recurrent policy network, $f_\pi$ indicates the recurrent policy network and $I_{t-1}$ denotes the restored face image at step $t-1$. The state $s_{t-1}$ is a vector, encoded with the current state, and $g$ refers to the cropping operation, which is applied to crop the corresponding patch $\hat{I}_{t-1}^{l_t}$ given action $l_t$.

Given attention patch $\hat{I}_{t-1}^{l_t}$, the local enhancement network $f_e$ is adopted for hallucination.
\begin{equation}
\hat{I}_{t}^{l_t} = f_{e}(\hat{I}_{t-1}^{l_t},I_{t-1};\theta_e).
\end{equation}

where $\theta_e$ is the parameter of the local enhancement network and $\hat{I}_{t}^{l_t}$ indicates the enhanced patch. After local enhancement, we replace $\hat{I}_{t-1}^{l_t}$ with $\hat{I}_{t}^{l_t}$. Specifically, the restored face image $I_{t}$ for the next step $t$ is produced by replacing the existing patch $\hat{I}_{t-1}^{l_t}$ with the enhanced patch $\hat{I}_{t}^{l_t}$. The overall coarse-to-fine face hallucination can be defined as:
\begin{equation}
\begin{cases}
I_0 = I_{lr}               &\\
I_t = f(I_{t-1};\theta)  & 1 \leq t \leq T, \\
I_{hr} = I_T
\end{cases}
\end{equation}
where $I_0$ is the original corrupted face image, $I_t$ indicates the enhanced full-sized facial image at each step $t$, $T$ is the maximal recurrent step, $\theta = [\theta_\pi;\theta_e]$ and $f = [f_\pi;f_e]$. $T$ is set to 18 according to our empirical analyses, which are presented in Section~\ref{sec:experiments}.

\begin{table*}[]
\centering
\footnotesize
\begin{floatrow}
\capbtabbox{
\begin{tabular}{cccccc}
\hline
                             & 1                         & 2                         & 3                         & 4                         & 5  \\ \hline
\multicolumn{1}{l|}{layer}   & \multicolumn{1}{l|}{conv} & \multicolumn{1}{l|}{conv} & \multicolumn{1}{l|}{conv} & \multicolumn{1}{l|}{conv} & fc \\ \hline
\multicolumn{1}{l|}{stride}  & \multicolumn{1}{l|}{2}    & \multicolumn{1}{l|}{2}    & \multicolumn{1}{l|}{2}    & \multicolumn{1}{l|}{2}    & -  \\ \hline
\multicolumn{1}{l|}{size}    & \multicolumn{1}{l|}{64}   & \multicolumn{1}{l|}{32}   & \multicolumn{1}{l|}{16}   & \multicolumn{1}{l|}{8}    & 64 \\ \hline
\multicolumn{1}{l|}{kernel}  & \multicolumn{1}{l|}{5}    & \multicolumn{1}{l|}{3}    & \multicolumn{1}{l|}{3}    & \multicolumn{1}{l|}{3}    & -  \\ \hline
\multicolumn{1}{l|}{channel} & \multicolumn{1}{l|}{8}    & \multicolumn{1}{l|}{8}    & \multicolumn{1}{l|}{16}   & \multicolumn{1}{l|}{32}   & 1  \\ \hline
\end{tabular}
}
{
 \caption{Detailed setting of each component in the feature extractor.}
 \label{table:feature_exactor}
}
\capbtabbox{

\begin{tabular}{cccccccc}
\hline
                             & 1                         & 2                         & 3                           & 4                         & 5                           & 6                         & 7    \\ \hline
\multicolumn{1}{l|}{layer}   & \multicolumn{1}{l|}{conv} & \multicolumn{1}{l|}{conv} & \multicolumn{1}{l|}{deconv} & \multicolumn{1}{l|}{conv} & \multicolumn{1}{l|}{deconv} & \multicolumn{1}{l|}{conv} & conv \\ \hline
\multicolumn{1}{l|}{stride}  & \multicolumn{1}{l|}{1}    & \multicolumn{1}{l|}{1}    & \multicolumn{1}{l|}{2}      & \multicolumn{1}{l|}{1}    & \multicolumn{1}{l|}{2}      & \multicolumn{1}{l|}{1}    & 1    \\ \hline
\multicolumn{1}{l|}{size}    & \multicolumn{1}{l|}{8}    & \multicolumn{1}{l|}{16}   & \multicolumn{1}{l|}{32}     & \multicolumn{1}{l|}{16}   & \multicolumn{1}{l|}{8}      & \multicolumn{1}{l|}{8}    & 8    \\ \hline
\multicolumn{1}{l|}{kernel}  & \multicolumn{1}{l|}{5}    & \multicolumn{1}{l|}{3}    & \multicolumn{1}{l|}{3}      & \multicolumn{1}{l|}{3}    & \multicolumn{1}{l|}{3}      & \multicolumn{1}{l|}{3}    & 5    \\ \hline
\multicolumn{1}{l|}{channel} & \multicolumn{1}{l|}{64}   & \multicolumn{1}{l|}{32}   & \multicolumn{1}{l|}{32}     & \multicolumn{1}{l|}{32}   & \multicolumn{1}{l|}{8}      & \multicolumn{1}{l|}{8}    & 1    \\ \hline
\end{tabular}
}{
 \caption{Detailed setting of each component in the local enhancement network.}
 \label{table:enhancementnet}
}
\end{floatrow}
\end{table*}

\subsection{Recurrent Policy Network}
The recurrent policy network is designed to cooperate with a recurrent neural network to optimize a time-sequence of attended regions for local enhancement. As illustrated in Fig.~\ref{fig:model_overview}, the Attention-FH framework is composed of a recurrent policy network and a local enhancement network. The recurrent policy network can be formulated as a decision-making process for optimal patch selection on time intervals. At each step, the policy network takes as input the concatenation feature vector~(i.e.,~the state), which contains the current enhanced image, the original corrupted image and the encoded historical actions, and learns to determine the optimal image patch to be enhanced at each time step. At the final time step $T$, a global delayed reward, which is measured in terms of the accumulated attended rate and the mean squared error (MSE) between the hallucinated image and the corresponding ground truth, is used to guide the training of the policy network. The agent learns to predict the most appropriate restoring route for different identities by maximizing the global delayed reward. 
% \yukai{The} messages are all encoded into a feature vector and represented as the state to the agent. \yukai{Then}, the policy network \yukai{picking and hallucinates patches by} such means recursively. At the last step $T$, a global delayed reward, which \yukai{consists of the} MSE distance and \yukai{the} attending rate, is given to the policy network. By maximizing the global delayed reward, the agent \yukai{learns} to predict the most appropriate restoring route for different \yukai{identities}.

\textbf{State.} \syk{To provide rich contextual information, the state of the agent is designed to contain the three following types of information. 1) The enhanced face $I_t$ from previous steps, which enables the agent to determine the patch that needs to be repaired at the next time step by fully sensing the rich contextual information~(e.g., the region that is still LR and needs to be enhanced). $I_t$ is represented as a global feature vector $v_{t}$ extracted from the output of a fully connected layer. 2) The original corrupted face image $I_0$, which is also encoded with a global feature vector $v_{0}$, as with the enhanced facial image. 3) The encoded history action vector $v_l$ obtained by forwarding all previous action vectors $\{h_0,h_1,...,h_{t-1}\}$ into the GRU network. The output hidden variable of the GRU thus encodes all previous action information and is denoted as $v_l$. We formulate the state $s_t$ with size of 512 $\times$ 1 to encode multiple contextual information $\{v_t, v_0, v_l\}$. In this way, the target of the agent is to determine the region cropping action~(including the location and the size) of the next attended local patch by considering the state $s_t$.
}

 % \yukai{The current} enhanced face image $I_t$. To \yukai{accurately} provide restored \yukai{conditions} to the agent, the $I_t$ is first \qingxing{extracted as a 64-d} feature vector $v_{t}$. 2) \yukai{The original} corrupted face image $I_0$. \yukai{To offer} rich contextual information, the $I_0$ is represented into a 64 $\times$ 1 feature vector $v_{0}$. The setting of our feature \yukai{extractor is summarized} in Table.~\ref{table:feature_exactor}. 3) \yukai{The history locations} $v_{l}$. In addition to \yukai{the} information mentioned above, we also \yukai{pass} the history locations $v_{l}$ to the agent. The history locations are obtained by encoding the latent variables of the recurrent network. Specifically, the latent variables of the Gated Recurrent Unit(GRU) is used as a reference for the agent. It is used to locate the attention region \yukai{with the information of all history locations}. The state $s_t$ at step $t$ is produced by concatenating the vectors $\{v_{t},v_0,v_{l}\}$. To this end, the agent can observe the current state $s_t = \{v_{t},v_0,v_{l}\}$. \yukai{The next task} is to predict the location and the size of attention region.
\begin{figure}[t]
    \centering
    \includegraphics[height=0.1\textheight]{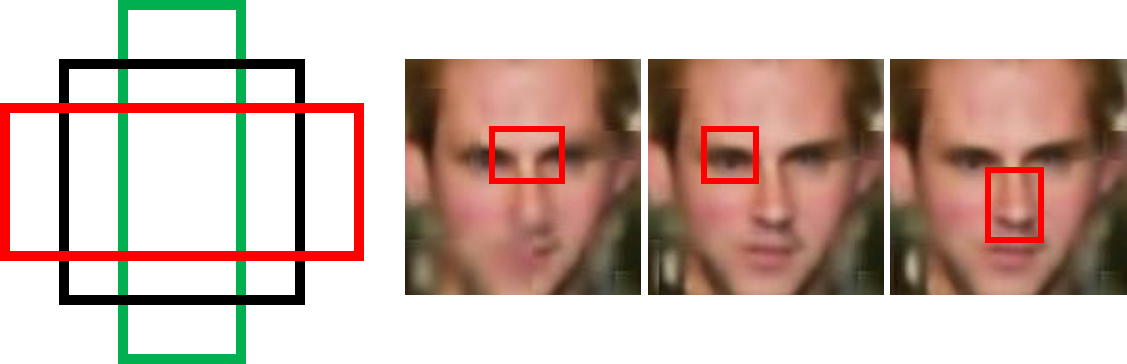}
    \caption{A demonstration of size-free attention. As the facial part has different sizes with particular identities, our method locates the region with a flexible attention mechanism.}
    \label{fig:ratio}
    \vspace{3mm}
\end{figure}
\textbf{Action.}
Given state $s_t$, the agent attempts to generate an action indication, which represents the next attended local region $\hat{I}_{t-1}^{l_t}$~(cropping from the last hallucinated image $\hat{I}_{t-1}$) for enhancement. Due to the large differences in the orientation and size of faces in in-the-wild cases, the use of a fixed-size attention bounding box to capture all facial components is not optimal. To this end, we propose a content-adaptive and size-free attention mechanism for better local region extraction. Specifically, to fully capture one facial component, the agent needs to predict the $(x,y,w,h)$ of a rectangular region at each time step, where $x$,$y$,$w$, and $h$, respectively, refers to the center coordinates and the width and height of the bounding box. Let $W$ and $H$ be the width and height of the target hallucinated facial image, and let $x,y = (x,y|1 \leq x \leq W,1\leq y \leq H)$. To reduce the search space, we give up the free search of the width and height and instead attempt to predict the ratio factor $ID_{ratio}$ and the scale factor $ID_{scale}$.  $ID_{ratio}$ is used to control the length-width ratio. Empirical candidates for $ID_{ratio}$ include $\{3:2,1:1,2:3\}$. As shown in Fig.~\ref{fig:ratio}, the bounding box is adapted w.r.t the ratio factor and is thus sufficiently flexible to handle various facial components. The scale factor $ID_{scale}$ is adopted to represent the overall size of the attention box. With the sample factors, the size of the attention box can be obtained by:
\begin{equation}
\begin{aligned}
L_h = &\mathrm{Z} * ID^{h}_{scale} \\
L_w = &L_h/ID_{ratio} *  ID^{w}_{scale}
\end{aligned}
\end{equation}

where $L_h$ and $L_w$ are the height and the width of the attention box, respectively. $\mathrm{Z}$ is a constant value that specifies the initial size of the attended region. Here, we set $\mathrm{Z}$ to 60. The action $l_t$ consists of $\{x,y,ID_{ratio},ID_{scale},\}$. The GRU takes the state $s_{t}$ as input and generates a 128 $\times$ 1 hidden vector $v_{c}$, which is fed to a fully connected layer to infer the action of the next step. We employ a tensor with the same size as the full image to ensure that the attended region can be accommodated.

\textbf{Reward.}
The reward function is designed to guide the training of the recurrent policy network for the optimization of a time sequence of attended patches for local enhancement. We consider two factors when designing our reward function. 1) The MSE between the hallucinated image after $T$ steps of local enhancement and the corresponding HR image. {\yukai{Specifically, let $I_{T}$ be the enhanced image at the last step $T$, and let $I_{s}$ be the image restored by the super-resolution generative adversarial network (SRGAN)~\cite{srgan}.} We first compute their MSEs with respect to the corresponding HR ground truth, denoted as $\mathrm{E}_T$ and $\mathrm{E}_{SRGAN}$. The first term of our reward function is defined as $\mathrm{E}_T-\mathrm{E}_{SRGAN}$. $\mathrm{E}_T$ measures the absolute peak signal-to-noise ratio (PSNR) of the restored image while the introduction of the second $\mathrm{E}_{SRGAN}$ replaces the reward function with the relative change in PSNR, which better reflects the evolution of each iteration and greatly enhances the stability of the model training. For instance, the PSNR of a restored image with simple details is usually much higher than that of an image with a relatively complex texture. By subtracting $\mathrm{E}_{SRGAN}$, the reward function can better reflect the incremental situation of model training and thus apply more accurate rewards and punishments. 2) The attention rate, which is introduced to indicate whether the attended region has covered the whole image. In detail, a tensor $T_0$ of the same size as the full image is employed. Initially, $T_0$ is set with all zeros, and once a region is visited, its corresponding value is set to $1$. We sum the tensor $T_0$ to $\mathrm{E_{v}}$ at the last step to reflect the attention ratio. The total reward function can be written as:

\begin{equation}
r_t = \begin{cases}
0     &t < T\\
\mathrm{E}_{T}-\mathrm{E}_{SRGAN} + \mathrm{E_{V}}   &t = T.
\end{cases}
\end{equation}

During training, we set the reward in a global manner, i.e., the reward is assigned after step $T$ is completed. The recurrent policy network is trained to maximize this reward via the REINFORCE algorithm~\cite{REINFORCE}.

\begin{figure}[t]
    \centering
    \includegraphics[width=0.85 \columnwidth]{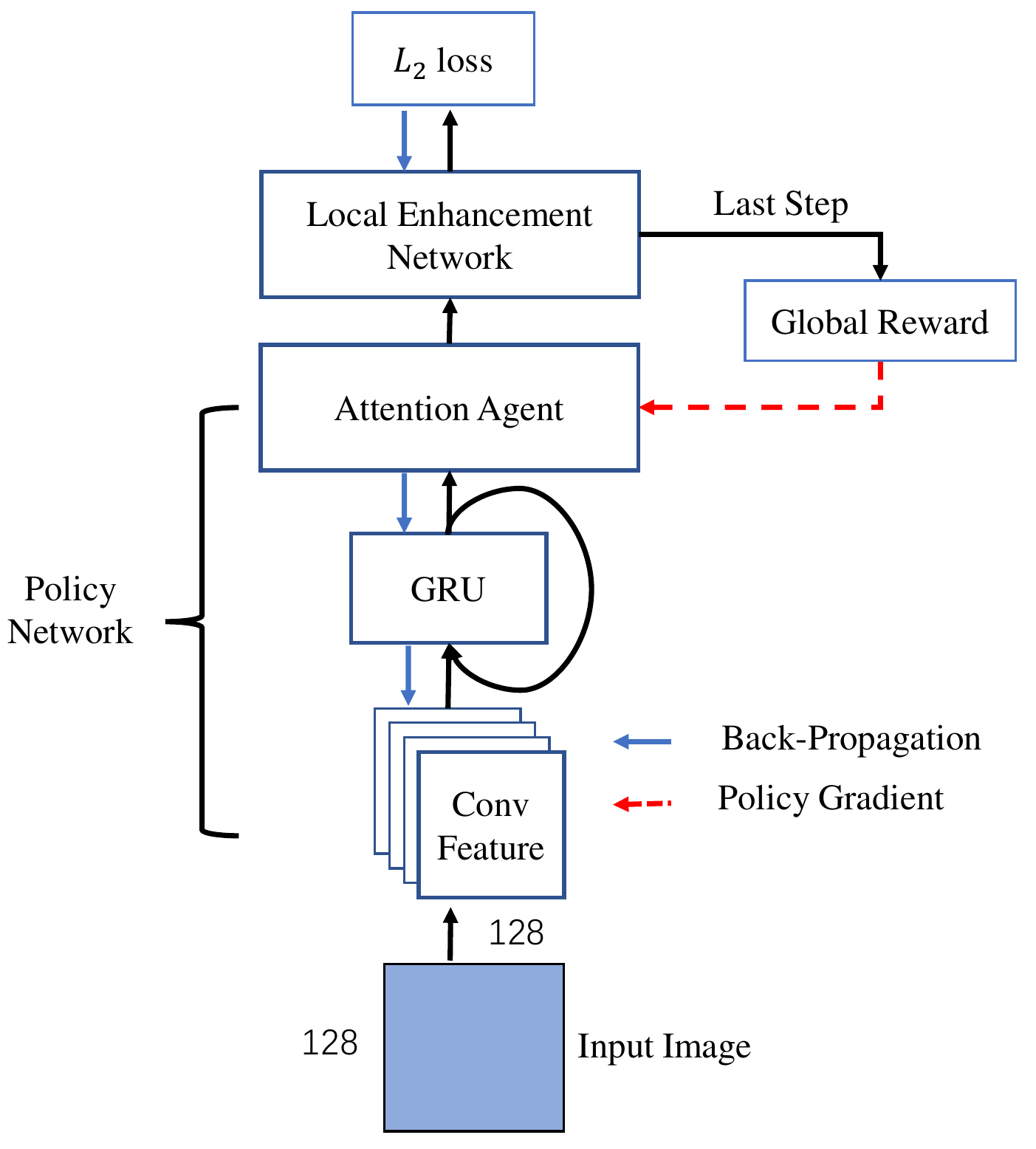}
    \caption{Detailed training procedure of our Attention-FH. The local enhancement network is optimized via the $L_2$ loss between the restored patch and the corresponding HR ground truth. After performing the last step, we calculate the global reward and pass the policy gradient to optimize the recurrent policy network.}
   
    \label{fig:train}
    \vspace{2mm}
\end{figure}

\begin{algorithm}[t]
\caption{Learning Algorithm of Attention-FH}
\label{alg:A}
\begin{algorithmic}[1]
\REQUIRE Training LR face images $I_{lr}$; HR face images $I_{hr}$; Initial actions $h_{0}$
\WHILE{t $=$ 1}
\STATE Represent $I_{lr}$, $I_{lr}$, $h_{0}$ as feature vectors $v_0$, $v_0$, $v^{0}_l$
\STATE Obtain state $s_0 := \{v_t, v_0, v^{0}_l\}$
\STATE Obtain actions $l_0 := f_\pi(s_{0};\theta_\pi)$
\STATE Crop out patch with respect to actions $\hat{I}_{0}^{l_0} := g(l_0,I_{0})$
\STATE Enhance the patch $\hat{I}_{1}^{l_0} := f_{e}(\hat{I}_{0}^{l_0},I_{0},I_{t},I_G;\theta_e)$
\STATE Replace ${I}_{0}^{l_0}$ with $\hat{I}_{1}^{l_0}$ to produce restored image $I_1$
\STATE Update $\theta_e$ via $\hat{I}_{0}^{l_0}$ and ground truth patch $\hat{I}_{gt}^{l_0}$: $\frac{\partial L_e}{\partial f_e(\theta_e,\hat{I}_{0}^{l_0}) }$
\ENDWHILE  

\WHILE{t $\le$ T}
\STATE Represent $I_{lr}$, $I_{lr}$ as feature vectors $v_t$, $v_0$
\STATE Forward all previous action vectors $\{h_0,h_1,...,h_{t-1}\}$ and obtain history actions $v_l$
\STATE Obtain state $s_t := \{v_t, v_0, v_l\}$
\STATE Obtain actions $l_t := f_\pi(s_{t};\theta_\pi)$
\STATE Crop out patch with respect to actions $\hat{I}_{t}^{l_t} := g(l_t,I_{t})$
\STATE Enhance the patch $\hat{I}_{t+1}^{l_t} := f_{e}(\hat{I}_{t}^{l_t},I_{0},I_{t},I_G;\theta_e)$
\STATE Replace ${I}_{t}^{l_t}$ with $\hat{I}_{t+1}^{l_t}$ to produce restored image $I_t$
\STATE Update $\theta_e$ via $\hat{I}_{t}^{l_t}$ and ground truth patch $\hat{I}_{gt}^{l_t}$: $\frac{\partial L_e}{\partial f_e(\theta_e,\hat{I}_{t}^{l_t}) }$
\IF{t $=$ T}
\STATE Calculate reward $r_t := \mathrm{E}_{T}-\mathrm{E}_{SRGAN} + \mathrm{E_{V}}$
\STATE Update $\theta_\pi$ via REINFORCE algorithm: $\frac{dr}{dl_t} := a *(r-b)*\frac{d\ln(f_\pi(x,l_t))}{dl_t}$
\ENDIF 
\ENDWHILE  
\end{algorithmic}
\end{algorithm}

\subsection{Local Enhancement Network}
Given an attended patch from the recurrent policy network, the local enhancement network $f_{\pi}$ is employed for hallucination. To provide comprehensive contextual information, $f_{\pi}$ takes the following three components as input: 1) The attended patch $\hat{I}_{t}^{l_t}$, which is represented by masking the outside area of the input image~(setting the value of the region outside the selected patch to zero while keeping the pixels inside intact). 2) The current enhanced facial image $I_t$~(with all previously hallucinated results pasted on), which provides global contextual information for the enhancement of the current patch. 3) The original corrupted image $I_0$, {\yukai{4) and the global context $I_G$, which is calculated by expanding $s_t$ to the dimension equal to the size of the input image $I_0$ by a fully connected layer followed by a reshaping operation~(i.e., $I_G$ is of the same shape of $I_0$).}} $\{\hat{I}_{t}^{l_t},I_t,I_0,I_G\}$ are concatenated and further fed into the local enhancement network. To achieve a trade-off between performance and efficiency, we adopt a reduced version of LapSRN as our local enhancement network. The simplified settings of LapSRN are listed in Table~\ref{table:enhancementnet}. We resize the three input components to the same size as the LR corrupted image without interpolation to improve the efficiency of our model. The local enhancement network is fully convolutional and is composed of 5 convolutional and 2 deconvolutional layers. The convolutional layers all have a stride of $1$, the kernels of the head and tail layers are of size $5*5$ and the kernels of the other layers are of size $3*3$. By incorporating two deconvolutional layers with stride $2$, the network is able to learn to reconstruct the resolution of the corrupted patch in a cascaded manner. At the end of the local enhancement network, the LR input image is upscaled to the target resolution. We crop out the attended region in the hallucinated result w.r.t its size and location. The cropped HR patch $\hat{I}_{t+1}^{l_t}$ is added to the accumulated enhanced result $I_t$. Finally, the residual between the attention patch $\hat{I}_{t}^{l_t}$ and the ground truth face image $I_{hr}$ can be estimated by the well-trained local enhancement network.

\subsection{Model Training}

We illustrate the training strategy of our Attention-FH in Fig.~\ref{fig:train}. The recurrent policy network, which learns to obtain the attentional patch by maximizing the reinforced reward, is shown in Fig.~\ref{fig:model_overview}(1). Fig.~\ref{fig:model_overview}(2) shows the local enhancement network, which learns to enhance the attended patch in an end-to-end mode by minimizing the MSE. 

In the training phase, the recurrent policy network is optimized by the REINFORCE algorithm~\cite{REINFORCE}, guided by the reward calculated at the end of sequential enhancement when the maximum time step $T$ is reached. The local enhancement network is optimized to minimize the $L_2$ distance between the restored patch and its corresponding HR ground truth. The supervised loss is calculated at each time step and can be minimized based on
backpropagation. After calculating the last step of attending patches for local enhancement, we obtain the global reward, which is leveraged to optimize the policy network. 

The whole training algorithm of our Attention-FH is illustrated in Alg.~\ref{alg:A}, which accords with the pipeline of our proposed framework shown in Fig.~\ref{fig:intro}.

\section{Experiments}
\label{sec:experiments}\label{sec:exper}
To demonstrate the advantages of our Attention-FH, we have conducted extensive experiments on multiple widely used benchmarks, i.e., \emph{CelebFaces Attributes Dataset}~\cite{liu2015deep}, \emph{Public Figures Face Dataset}~\cite{kumar2009attribute}, \emph{Labeled Faces in the Wild Dataset}~\cite{lfw-funneled}, \emph{Surveillance Cameras Face Dataset}~\cite{grgic2011scface} and \emph{BioID Face Dataset}~\cite{bioid}. We first briefly introduce the evaluation datasets, the corresponding evaluation protocols and the implementation details. Then, we perform comprehensive comparisons to verify the superiority of our Attention-FH over all the compared state-of-the-art approaches. Note that because the degradation types of face hallucination are complex, we have employed several down-sampling factors to evaluate our Attention-FH under various challenging conditions. Finally, we have performed detailed ablation studies to demonstrate the contribution of each component within our Attention-FH.

\begin{table*}[]
\centering
\footnotesize
\begin{tabular}{|c|c|c|c|c|c|c|c|c|}
\hline
\multicolumn{1}{|c|}{\multirow{2}{*}{Dataset}} & \multicolumn{1}{c|}{\multirow{2}{*}{Scale}} & \multicolumn{1}{c|}{\multirow{2}{*}{Bicubic}} & \multicolumn{2}{c|}{General SR}   & \multicolumn{2}{c|}{Face Hallucination} & \multicolumn{1}{c|}{GAN Based} & \multicolumn{1}{c|}{\multirow{2}{*}{Our}} \\ \cline{4-8}
\multicolumn{1}{|c|}{}                         & \multicolumn{1}{c|}{}                       & \multicolumn{1}{c|}{}                         & SRCNN & \multicolumn{1}{c|}{VDSR}  & BiCNN & \multicolumn{1}{c|}{GLN} & \multicolumn{1}{c|}{SRGAN}     & \multicolumn{1}{c|}{}                     \\ \hline \hline
\multirow{3}{*}{PubFig} & $\times$4    & 24.76   & 25.21 & \underline{28.05}    & 24.35 & 26.12 & 27.44 & \textbf{28.87} \\
                        & $\times$8    & 20.75   & 21.31 & 21.94    & 20.69 & 21.33 & \underline{23.45} & \textbf{23.59} \\
                        & $\times$16   & 17.89   & 18.48 & 19.28    & 18.13 & 18.73 & \underline{20.25} & \textbf{20.31} \\ \hline \hline
\multirow{3}{*}{CelebA} & $\times$4    & 25.76   & 26.01 & 28.38    & 24.93 & \underline{29.92} & 28.17 & \textbf{30.58} \\
                        & $\times$8    & 21.84   & 22.64 & 24.46    & 21.32 & \underline{25.48} & 24.37 & \textbf{26.14} \\
                        & $\times$16   & 18.78   & 19.80 & 20.07    & 19.01 & 21.20 & \underline{21.99} & \textbf{22.63} \\ \hline \hline
\multirow{3}{*}{SCface} & $\times$4    & 26.15   & 26.54 & \underline{31.59}    & 26.04 & 29.84 & 30.10 & \textbf{34.01} \\
                        & $\times$8    & 20.83   & 21.68 & 24.12    & 20.67 & 24.10 & \underline{24.72} & \textbf{26.04} \\
                        & $\times$16   & 17.15   & 18.45 & 20.32    & 17.21 & 18.28 & \underline{20.63} & \textbf{22.19} \\ \hline \hline
\multirow{3}{*}{BioID}  & $\times$4    & 24.59   & 25.71 & \underline{29.38}    & 23.16 & 26.71 & 28.16 & \textbf{33.38} \\
                        & $\times$8    & 20.24   & 21.85 & \underline{23.95}    & 20.14 & 22.03 & 23.23 & \textbf{27.81} \\
                        & $\times$16   & 17.15   & 18.45 & 19.41    & 17.11 & 19.85 & \underline{21.59} & \textbf{23.48} \\ \hline \hline
\multirow{3}{*}{LFW}  & $\times$4    & 26.79   & 28.94 & \underline{32.11}    & 26.60 & 30.34 & 31.41 & \textbf{32.93} \\
                        & $\times$8    & 21.92   & 23.92 & 24.12    & 22.62 & 24.51 & \underline{25.49} & \textbf{27.81} \\
                        & $\times$16   & 19.95   & 21.34 & 22.40    & 20.82 & 22.44 & \underline{23.01} & \textbf{23.13} \\ \hline
\end{tabular}
\caption{Quantitative comparison of our model and the competing methods in terms of the PSNR index. We use \textbf{bold face} and \underline{underline} to indicate the first and second place in each dataset.}
\label{tab:psnr}
\end{table*}

\begin{table*}[]
\centering
\footnotesize
\begin{tabular}{|c|c|c|c|c|c|c|c|c|}
\hline
\multicolumn{1}{|c|}{\multirow{2}{*}{Dataset}} & \multicolumn{1}{c|}{\multirow{2}{*}{Scale}} & \multicolumn{1}{c|}{\multirow{2}{*}{Bicubic}} & \multicolumn{2}{c|}{General SR}   & \multicolumn{2}{c|}{Face Hallucination} & \multicolumn{1}{c|}{GAN Based} & \multicolumn{1}{c|}{\multirow{2}{*}{Our}} \\ \cline{4-8}
\multicolumn{1}{|c|}{}                         & \multicolumn{1}{c|}{}                       & \multicolumn{1}{c|}{}                         & SRCNN & \multicolumn{1}{c|}{VDSR}  & BiCNN & \multicolumn{1}{c|}{GLN} & \multicolumn{1}{c|}{SRGAN}     & \multicolumn{1}{c|}{}                     \\ \hline \hline
\multirow{3}{*}{PubFig} & $\times$4    & 0.7600   & 0.7708 & \underline{0.8262}  & 0.7167 & 0.7793 & 0.8197 & \textbf{0.8587} \\
                        & $\times$8    & 0.5782   & 0.5819 & 0.5822    & 0.5651 & 0.5375 & \underline{ 0.6747} & \textbf{0.6805} \\
                        & $\times$16   & 0.4581   & 0.4584 & 0.4761    & 0.4660 & 0.4203 & \underline{0.5384} & \textbf{0.5394} \\ \hline \hline
\multirow{3}{*}{CelebA} & $\times$4    & 0.7672   & 0.7717 & 0.8115    & 0.7663 & \underline{0.8587} & 0.8214 & \textbf{0.8711} \\
                        & $\times$8    & 0.6107   & 0.6228 & 0.6747    & 0.6093 & \underline{0.7141} & 0.6749 & \textbf{0.7441} \\
                        & $\times$16   & 0.5016   & 0.4995 & 0.5042    & 0.5019 & 0.5307 & \underline{0.5946} & \textbf{0.6316} \\ \hline \hline
\multirow{3}{*}{SCface} & $\times$4    & 0.8367   & 0.8435 & \underline{0.9036}    & 0.8396 & 0.8848 & 0.8907 & \textbf{0.9356} \\
                        & $\times$8    & 0.6303   & 0.6493 & 0.7009    & 0.6291 & 0.7132 & \underline{0.7567} & \textbf{0.7983} \\
                        & $\times$16   & 0.4821   & 0.5018 & 0.5355    & 0.4823 & 0.4962 & \underline{0.6175} & \textbf{0.6835} \\ \hline \hline
\multirow{3}{*}{BioID}  & $\times$4    & 0.8056   & 0.8244 & \underline{0.8618}    & 0.6719 & 0.8333 & 0.8600 & \textbf{0.9418} \\
                        & $\times$8    & 0.6189   & 0.6196 & 0.6676    & 0.5657 & 0.5996 & \underline{0.6806} & \textbf{0.8568} \\
                        & $\times$16   & 0.4821   & 0.5018 & 0.4908    & 0.4572 & 0.5305 & \underline{0.6546} & \textbf{0.7376} \\ \hline \hline
\multirow{3}{*}{LFW}    & $\times$4    & 0.8469   & 0.8686 & 0.8917    & 0.8329 & 0.8922 & \underline{0.9247} & \textbf{0.9418} \\
                        & $\times$8    & 0.6712   & 0.6927 & 0.7031    & 0.6801 & 0.7109 & \underline{0.7314} & \textbf{0.8568} \\
                        & $\times$16   & 0.5617   & 0.5742 & 0.5879    & 0.5838 & 0.6132 & \underline{0.6449} & \textbf{0.6542} \\ \hline

\end{tabular}
\caption{Quantitative comparison of our model and the competing methods in terms of the SSIM index. We use \textbf{bold face} and \underline{underline} to indicate the first and second place in each dataset.}
\label{tab:ssim}
\end{table*}

\begin{table*}[]
\centering
\footnotesize
\begin{tabular}{|c|c|c|c|c|c|c|c|c|}
\hline
\multicolumn{1}{|c|}{\multirow{2}{*}{Dataset}} & \multicolumn{1}{c|}{\multirow{2}{*}{Scale}} & \multicolumn{1}{c|}{\multirow{2}{*}{Bicubic}} & \multicolumn{2}{c|}{General SR}   & \multicolumn{2}{c|}{Face Hallucination} & \multicolumn{1}{c|}{GAN Based} & \multicolumn{1}{c|}{\multirow{2}{*}{Our}} \\ \cline{4-8}
\multicolumn{1}{|c|}{}                         & \multicolumn{1}{c|}{}                       & \multicolumn{1}{c|}{}                         & SRCNN & \multicolumn{1}{c|}{VDSR}  & BiCNN & \multicolumn{1}{c|}{GLN} & \multicolumn{1}{c|}{SRGAN}     & \multicolumn{1}{c|}{}                     \\ \hline \hline
\multirow{3}{*}{PubFig} & $\times$4    & 0.8454   & 0.8581 & \underline{0.8942}    & 0.8748 & 0.8748 & 0.8892 & \textbf{0.9099} \\
                        & $\times$8    & 0.7337   & 0.7604 & 0.7587    & 0.7669 & 0.7669 & \underline{0.8081} & \textbf{0.8121} \\
                        & $\times$16   & 0.6360   & 0.7872 & 0.7207    & 0.7151 & 0.7151 & \underline{0.7297} & \textbf{0.7346} \\ \hline \hline
\multirow{3}{*}{CelebA} & $\times$4    & 0.8565   & 0.8634 & 0.8890    & 0.9148 & \underline{0.9148} & 0.8956 & \textbf{0.9211} \\
                        & $\times$8    & 0.7514   & 0.7762 & 0.8077    & 0.8290 & \underline{0.8290} & 0.8066 & \textbf{0.8410} \\
                        & $\times$16   & 0.6511   & 0.7098 & 0.6980    & 0.7331 & 0.7331 & \underline{0.7552} & \textbf{0.7628} \\ \hline \hline
\multirow{3}{*}{SCface} & $\times$4    & 0.8813   & 0.8906 & \underline{0.9321}    & 0.9185 & 0.9185 & 0.9205 & \textbf{0.9513} \\
                        & $\times$8    & 0.7518   & 0.7893 & 0.8195    & 0.8318 & 0.8318 & \underline{0.8437} & \textbf{0.8675} \\
                        & $\times$16   & 0.6487   & 0.7055 & 0.7236    & 0.6955 & 0.6955 & \underline{0.7586} & \textbf{0.7977} \\ \hline \hline
\multirow{3}{*}{BioID}  & $\times$4    & 0.8538   & 0.8767 & \underline{0.9066}    & 0.8926 & 0.8926 & 0.9012 & \textbf{0.9571} \\
                        & $\times$8    & 0.7405   & 0.7800 & 0.7977    & 0.7803 & 0.7803 & \underline{0.8162} & \textbf{0.9072} \\
                        & $\times$16   & 0.6487   & 0.7055 & 0.7104    & 0.7379 & 0.7379 & \underline{0.7849} & \textbf{0.8336} \\ \hline \hline
\multirow{3}{*}{LFW}    & $\times$4    & 0.8947   & 0.9069 & 0.9300    & 0.8982 & 0.9151 & \underline{0.9384}     & \textbf{0.9571} \\
                        & $\times$8    & 0.7824   & 0.8314 & 0.8391    & 0.7903 & \underline{0.8405} & 0.8278     & \textbf{0.9072} \\
                        & $\times$16   & 0.6771   & 0.7454 & 0.7595    & 0.7243 & 0.7788 & \underline{0.7722} & \textbf{0.7824} \\ \hline
\end{tabular}
\caption{Quantitative comparison of our model and the competing methods in terms of the FSIM index. We use \textbf{bold face} and \underline{underline} to indicate the first and second place in each dataset.}
\label{tab:fsim}
\end{table*}

\subsection{Datasets and Evaluation Protocols}
{\yukai{ We employ the following seven public datasets under various domains for a comprehensive evaluation to validate the robustness of our Attention-FH to in-the-wild faces.}}
\begin{itemize}
\item \emph{CelebA}~\cite{liu2015deep} is a large-scale dataset that contains 202,599 in-the-wild face images with 10,177 identities. Following the settings in~\cite{liu2015deep}, we adopt 188,311 images for training and use the remaining 14,288 for testing.
\item \emph{SCface}~\cite{grgic2011scface} consists of 4,160 images with 130 identities collected under an uncontrolled environment using five video surveillance cameras in various situations. As video surveillance is one of the main application scenarios for face hallucination, this dataset can be used to evaluate the compared methods from a practical perspective. We utilize 2,405 images for training and the rest for testing.
\item \emph{BioID}~\cite{bioid} is a public dataset with 1,521 gray face images, all taken in a laboratory from a frontal view. We use 1,028 images for training and the remaining 493 images for testing.
\item \emph{PubFig}~\cite{kumar2009attribute} is a large dataset with 58,797 real-world face images collected from the Web. In our experiment, 11,041 images are utilized for training, and the remaining 6,425 images are used for testing.
\item \emph{LFW}~\cite{lfw-funneled} contains 13,233 in-the-wild face images with 5,749 identities. We split the dataset w.r.t the partitions mentioned in~\cite{lfw-funneled}.
\item \yukai{\emph{Multi-PIE}~\cite{multipie} contains images of 337 subjects captured from complicated perspectives and under complex illumination conditions in four different sessions. We use 126,093 images for training and 31,524 images for testing.}
\item \yukai{\emph{Extended Yale-B}~\cite{yaleb} is a large face dataset that contains 16,128 images of 28 identities under 9 poses and 64 illumination conditions. We randomly choose 12,908 images for training and 3,220 images for evaluation.}

\end{itemize}

In terms of the evaluation metrics, we exploit PSNR and structural similarity (SSIM) to measure the performance of the compared methods. For a comprehensive evaluation, the feature similarity is also compared by adopting FSIM~\cite{zhang2011fsim}.

\subsection{Implementation Details}
%图像前处理（包括crop，normlization，align，down-sampling,）
%网络参数（recurrent step T，optimization strategy，learning rate，momentum term，size-free attention setting}
\syk{All the datasets are first aligned with two points of the eye regions by CFSS~\cite{Zhu2015CFSS}. Then, we simply crop the images to a size of 160 $\times$ 120 by prefetching the centric region, except for the LFW dataset, the images of which are cropped to 128$\times$128.} To ensure a fair comparison, all the methods are trained on only the corresponding training set, without using the other datasets for pre-training. We evaluate our method with scaling factors of $\times$4, $\times$8 and $\times$16 to model different types of situations. In addition, we also normalize the input images into $[-1,1]$. The recurrent time step $T$ of the policy network is set to 18 to achieve a trade-off between efficiency and accuracy. The setting of the recurrent time step $T$ is also investigated in Section~\ref{sec:ablation}. Our Attention-FH is trained using ADAM gradient descent~\cite{adam} with a base learning rate of $3 \times 10^{-4}$, a weight decay of $1e^{-7}$, and a momentum term of $0.5$. The training batch size is 16. Considering the absolute free attention region can lead to unstable performance, we impose some empirical constraints on the size-free attention mechanism, i.e., the length and width of the attention region are customized with respect to the ScaleID and RatioID, which are evaluated in Section~\ref{sec:ablation}.

\subsection{Competing Methods}
We compare our method with several state-of-the-art methods, including SRCNN~\cite{dong2014learning}, VDSR~\cite{vdsr}, SFH~\cite{yang2013structured}, BiCNN~\cite{bicnn}, GLN~\cite{gln}, and SRGAN~\cite{srgan}. These methods can be categorized into three groups: (i) general image super-resolution: SRCNN and VDSR; (ii) face hallucination: SFH, BiCNN and GLN; and (iii) generative adversarial learning: SRGAN. The first and second types are commonly applied to address regular image and face image restoration, respectively, while the third one is widely used in image generation and has achieved impressive results.

%%%%%%%%%%%%%%%%%%%%%%%%%%%%%%%%%%%%%%Fig for SCFace x8%%%%%%%%%%%%%%%%%%%%%%%%%%%%%%%%%%%%%%%%%%%%%%%%%%%%%%%%%%%%%%%%%
\begin{figure*}[]
\begin{minipage}{0.08\linewidth}
\centerline{\includegraphics[width=1\textwidth]{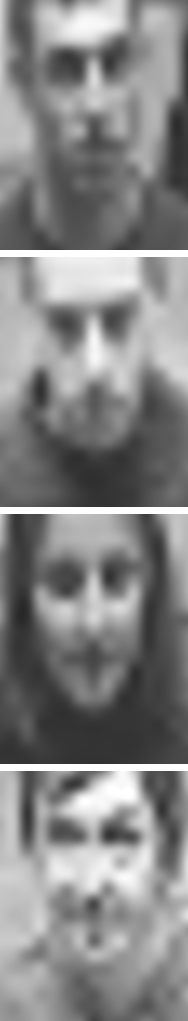}}
\centerline{Bicubic}
\end{minipage}
\!
\begin{minipage}{0.08\linewidth}
\centerline{\includegraphics[width=1\textwidth]{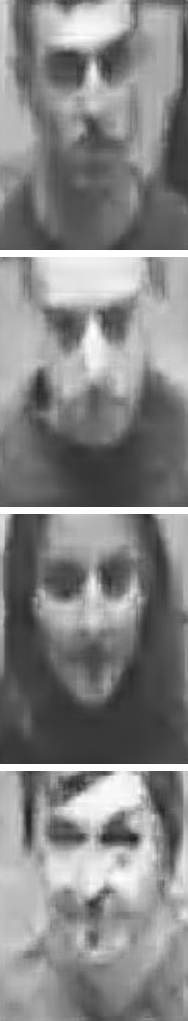}}
\centerline{SRCNN}
\end{minipage}
\!
\begin{minipage}{0.08\linewidth}
\centerline{\includegraphics[width=1\textwidth]{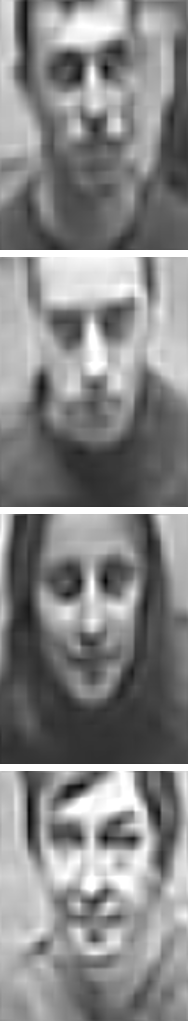}}
\centerline{VDSR}
\end{minipage}
\!
\begin{minipage}{0.08\linewidth}
\centerline{\includegraphics[width=1\textwidth]{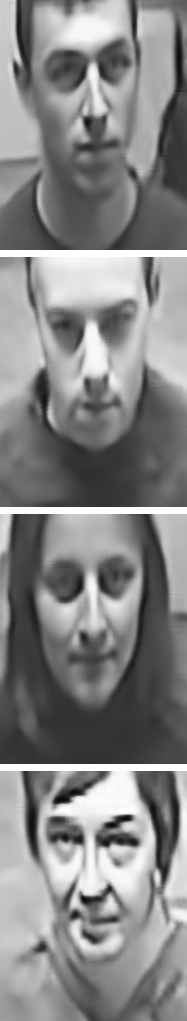}}
\centerline{Our}
\end{minipage}
\!
\begin{minipage}{0.08\linewidth}
\centerline{\includegraphics[width=1\textwidth]{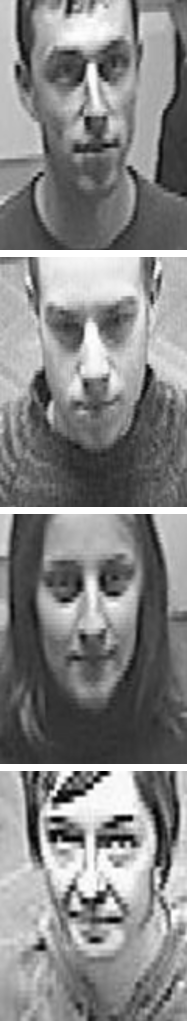}}
\centerline{Original}
\end{minipage}
\,
\begin{minipage}{0.08\linewidth}
\centerline{\includegraphics[width=1\textwidth]{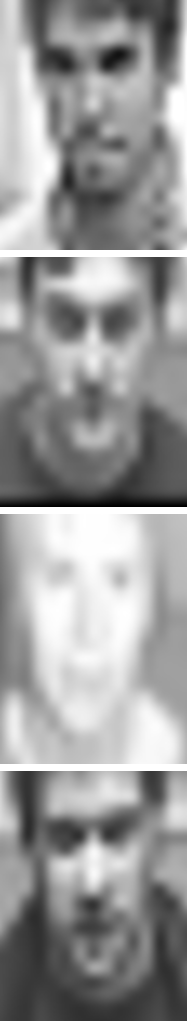}}
\centerline{Bicubic}
\end{minipage}
\!
\begin{minipage}{0.08\linewidth}
\centerline{\includegraphics[width=1\textwidth]{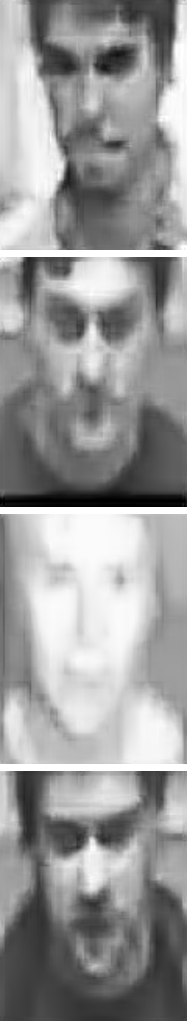}}
\centerline{SRCNN}
\end{minipage}
\!
\begin{minipage}{0.08\linewidth}
\centerline{\includegraphics[width=1\textwidth]{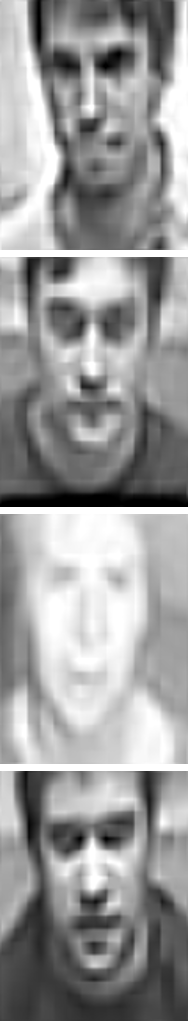}}
\centerline{VDSR}
\end{minipage}
\!
\begin{minipage}{0.08\linewidth}
\centerline{\includegraphics[width=1\textwidth]{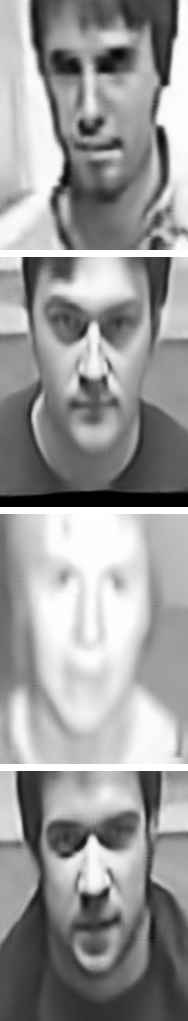}}
\centerline{Our}
\end{minipage}
\!
\begin{minipage}{0.08\linewidth}
\centerline{\includegraphics[width=1\textwidth]{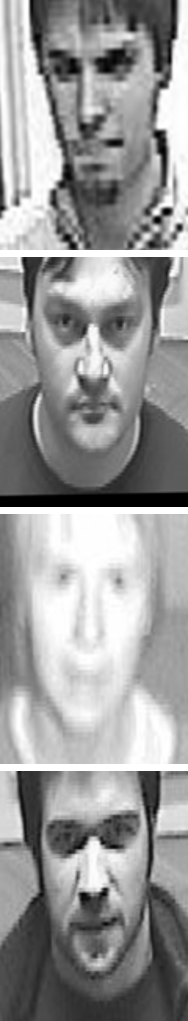}}
\centerline{Original}
\end{minipage}
\caption{Qualitative comparisons on the SCface~\cite{grgic2011scface} dataset with a scaling factor of 8. The testing faces are captured by surveillance cameras, similarly to practical scenarios. This figure is best viewed by zooming in on the electronic version.}
\label{fig:scface_x8}
\vspace{3mm}
\end{figure*}

\subsection{Quantitative and Qualitative Comparisons}
As illustrated in Tab.~\ref{tab:psnr},~\ref{tab:ssim} and~\ref{tab:fsim}, our Attention-FH consistently outperforms all the compared state-of-the-art methods, with clear margins in terms of all evaluation metrics. Attention-FH outperforms the best of the competing methods with 2.42 dB, 1.32 dB, and 1.56 dB on the SCface dataset with respect to the PSNR index, respectively. Moreover, our Attention-FH surpasses all the competing methods by large margins on all datasets when the scaling factor is small (e.g., 4). These results confirm the significant superiority of our Attention-FH.  
Note that we do not present the quantitative results of SFH~\cite{yang2013structured}, which relies heavily on face alignment and thus may fail to handle some testing images.

The visual comparison on the SCface dataset is presented in Fig.~\ref{fig:scface_x8}. Since the SCface dataset is similar to a real-world scenario, the dataset can be employed to explicitly validate the performance of each hallucination approach in terms of practicability. As shown in Fig.~\ref{fig:scface_x8}, regular super-resolution methods produce hallucinated facial images with blurry predictions. By contrast, our Attention-FH can generate faces with well-maintained facial structure. This result demonstrates that our Attention-FH is capable of deblurring and anti-aliasing facial images to preserve the structural information. 

The qualitative results shown in Fig.~\ref{fig:pubfig_x8},~\ref{fig:pubfig_x4} and~\ref{fig:pubfig_x16} demonstrate that our Attention-FH achieves significant improvements in restoration quality compared with all the competing methods. In addition, the attention mechanism also benefits our Attention-FH when addressing variation in pose, illumination and facial appearance. As depicted in Fig.~\ref{fig:pubfig_x8}, the facial expression of the woman in the third row is a `smile' with her mouth opened, and the latter woman has a `smile' with her mouth closed. Our Attention-FH outperforms all the competing methods in successfully addressing these two case with corrupted inputs. Furthermore, our Attention-FH can even hallucinate the man in the eighth row with `glasses', which is extremely challenging for all the compared state-of-the-art approaches. As demonstrated in Fig.\ref{fig:pubfig_x16}, our Attention-FH can recover naturally acceptable facial images even after substantial information has been lost by downsampling.

\yukai{ To further verify the effectiveness of our model, we also compare our proposed Attention-FH with some methods~\cite{edsr,pixelsr,image_transformer} proposed after the conference version of this paper. % implement recently proposed approaches for comparison~\cite{edsr,pixelsr,image_transformer}.
Pixel-SR~\cite{pixelsr} and Image Transformer~\cite{image_transformer} exhibit good performance in general object hallucination, and enhanced deep super-resolution network (EDSR) is famous for generating clear structures in general image SR. Since Pixel-SR and Image Transformer aim to hallucinate small-size objects (e.g., 32 $\times$ 32 pixels), the target resolution of face hallucination (e.g., 128 $\times$ 128 pixels) may lead to GPU memory explosion. Hence, following the official implementation, we conduct patch-based learning and combine the restored patches into a full image for evaluation. As shown in Tab.~\ref{table:deeper_comparsion}, our model produces better results than the other methods, except for EDSR. Nevertheless, Attention-FH is sufficiently flexible to incorporate EDSR as the local enhancement network to improve performance. We reduce the number of recurrent steps to 4 and implement EDSR for local enhancement. ``Our-EDSR'' achieves superior performance to EDSR as we expected, which well illustrates the flexibility and effectiveness of the proposed model. Furthermore, we average the results of three model snapshots with different iterations~(after convergence) of ``Our-EDSR'' and ``EDSR'' for comparison. As illustrated in Tab.~\ref{table:deeper_comparsion}, ``Our-EDSR ensemble'' achieves superior performance.
}

\yukai{Additionally, we compare Attention-FH on general image super-resolution with state-of-the-art image SR approaches~(i.e., VDSR, LapSRN, MemNet~\cite{memnet} and IDN~\cite{idn}). We follow the training scheme of IDN~\cite{idn} and conduct experiments on Set14~\cite{set14} to demonstrate the performance of our method on arbitrary domain images. With the overall parameter number fixed, we adaptively decrease the recurrent step and build up the local enhancement network for better image SR. As shown in Tab.~\ref{table:general_sr}, our method consistently outperforms all the compared methods on the $\times$2, $\times$3 and $\times$4 settings. Though Attention-FH is indeed capable of restoring the general image well, our model is still a face hallucination framework. By incorporating recurrent attention mechanism, Attention-FH is specialized in low-resolution faces and achieves much greater advantages in the task of face hallucination.
}

\begin{table}[]
\footnotesize
\centering
\begin{tabular}{|c|c|c|}
\hline
Algorithm         & Multi-PIE $\times$8 & Yale-B $\times$8   \\ \hline \hline
Bicubic           & 21.11  & 24.67  \\ \hline
SRCNN & 24.53  & 25.69   \\ \hline
VDSR & 26.12  & 26.57   \\ \hline
GLN & 28.17 & 25.98  \\ \hline 
SRGAN & 28.26 & 27.32  \\ \hline \hline
Our & 27.81  & 28.89 \\ \hline
\end{tabular}
\caption{Comparison on Multi-PIE~\cite{multipie} and Extended Yale-B~\cite{yaleb}.}
\label{table:comparison_on_multipie}
\end{table}

\begin{table}[]
\footnotesize
\centering
\begin{tabular}{|c|c|c|}
\hline
Algorithm         & LFW $\times$8 & LFW $\times$16   \\ \hline \hline
Pixel-SR           & 21.11  & 20.47  \\ \hline
Image Transformer & 24.53  & 22.22   \\ \hline
DRCN & 26.43  & 22.39   \\ \hline
EDSR & 28.17 & 24.10  \\ \hline 
EDSR ensemble & 28.26 & 24.21  \\ \hline \hline
Our & 27.81  & 23.13 \\ \hline
Our-EDSR   & 28.25  & 24.23  \\ \hline
Our-EDSR ensemble  & 28.30  & 24.28  \\ \hline
\end{tabular}
\caption{Comparison of our proposed model with deeper inferences.}
\label{table:deeper_comparsion}
\end{table}

\begin{table}[]
\footnotesize
\centering
\begin{tabular}{|c|c|c|c|}
\hline
Algorithm      & $\times$2 & $\times$3 & $\times$4   \\ \hline \hline
Bicubic           & 30.24  & 27.55 & 26.00 \\ \hline
VDSR & 33.03  & 29.77 & 28.01  \\ \hline
LapSRN & 32.99  & 29.79 & 28.09  \\ \hline
MemNet~\cite{memnet} & 33.28 & 30.00 & 28.26 \\ \hline 
IDN~\cite{idn} & 33.30 & 29.99 & 28.31  \\ \hline \hline
Our & 33.34  & 33.07 & 28.33\\ \hline
\end{tabular}
\caption{Comparison on general image super-resolution.}
\label{table:general_sr}
\end{table}

%%%%%%%%%%%%%%%%%%%%%%%%%%%%%%%%%%%%%%Fig for pubfig x8%%%%%%%%%%%%%%%%%%%%%%%%%%%%%%%%%%%%%%%%%%%%%%%%%%%%%%%%%%%%%%%%%
\begin{figure*}[]
\begin{minipage}{0.1\linewidth}
\centerline{\includegraphics[width=1\textwidth]{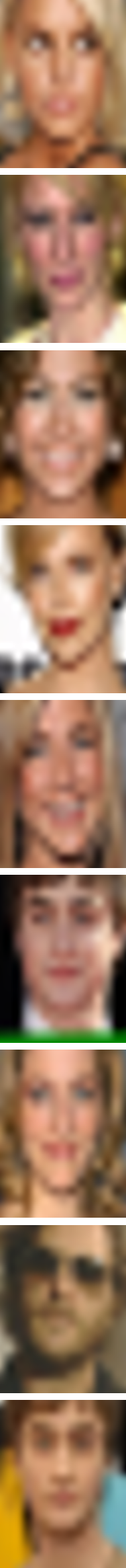}}
\centerline{Bicubic}
\end{minipage}
\!
\begin{minipage}{0.1\linewidth}
\centerline{\includegraphics[width=1\textwidth]{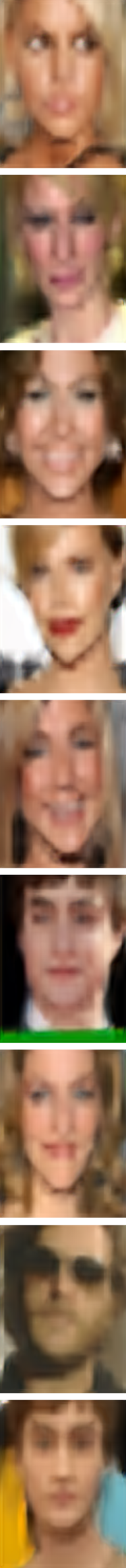}}
\centerline{SRCNN}
\end{minipage}
\!
\begin{minipage}{0.1\linewidth}
\centerline{\includegraphics[width=1\textwidth]{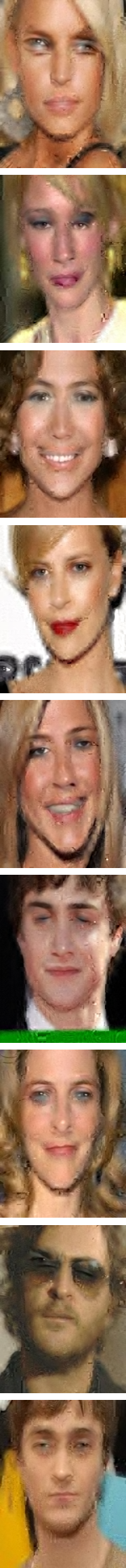}}
\centerline{SFH}
\end{minipage}
\!
\begin{minipage}{0.1\linewidth}
\centerline{\includegraphics[width=1\textwidth]{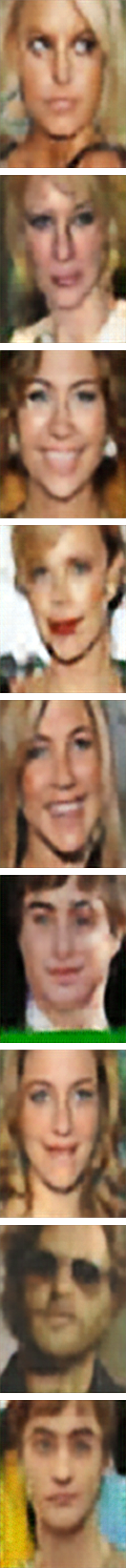}}
\centerline{GLN}
\end{minipage}
\!
\begin{minipage}{0.1\linewidth}
\centerline{\includegraphics[width=1\textwidth]{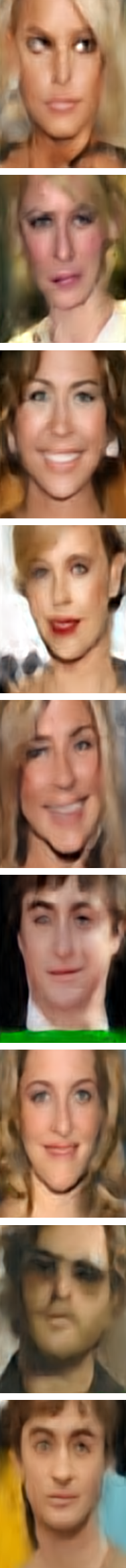}}
\centerline{VDSR}
\end{minipage}
\!
\begin{minipage}{0.1\linewidth}
\centerline{\includegraphics[width=1\textwidth]{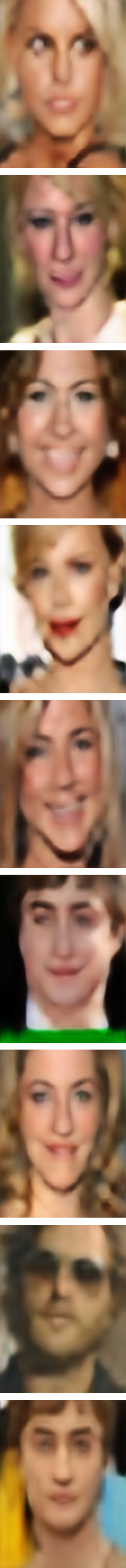}}
\centerline{SRGAN}
\end{minipage}
\!
\begin{minipage}{0.1\linewidth}
\centerline{\includegraphics[width=1\textwidth]{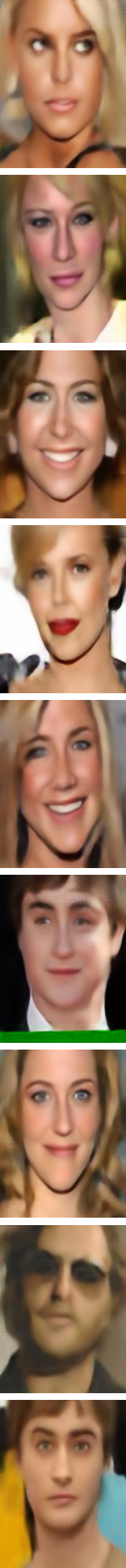}}
\centerline{Our}
\end{minipage}
\!
\begin{minipage}{0.1\linewidth}
\centerline{\includegraphics[width=1\textwidth]{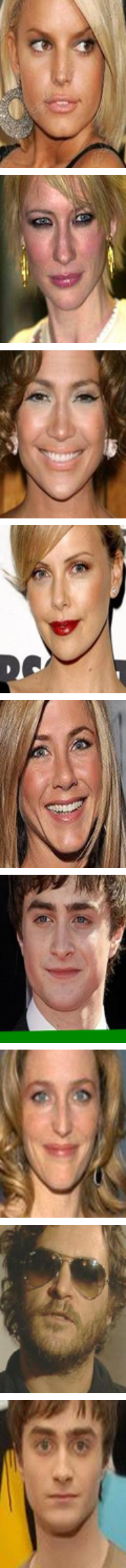}}
\centerline{Original}
\end{minipage}
\caption{Qualitative comparison on the PubFig~\cite{kumar2009attribute} dataset with a scaling factor of 8. The image is best viewed by zooming in on the electronic version.}
\label{fig:pubfig_x8}
\end{figure*}

%%%%%%%%%%%%%%%%%%%%%%%%%%%%%%%%%%%%%%Fig for pubfig x4%%%%%%%%%%%%%%%%%%%%%%%%%%%%%%%%%%%%%%%%%%%%%%%%%%%%%%%%%%%%%%%%%
\begin{figure*}[]
\begin{minipage}{0.1\linewidth}
\centerline{\includegraphics[width=1\textwidth]{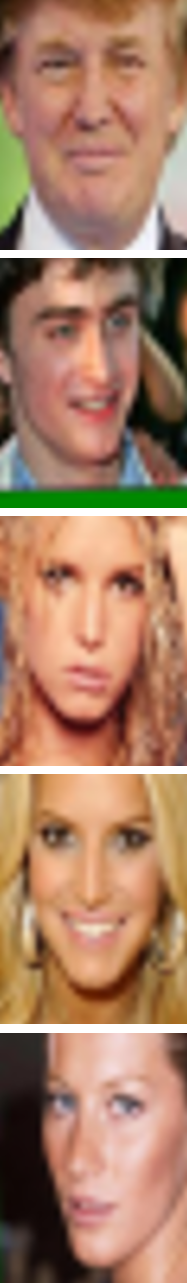}}
\centerline{Bicubic}
\end{minipage}
\!
\begin{minipage}{0.1\linewidth}
\centerline{\includegraphics[width=1\textwidth]{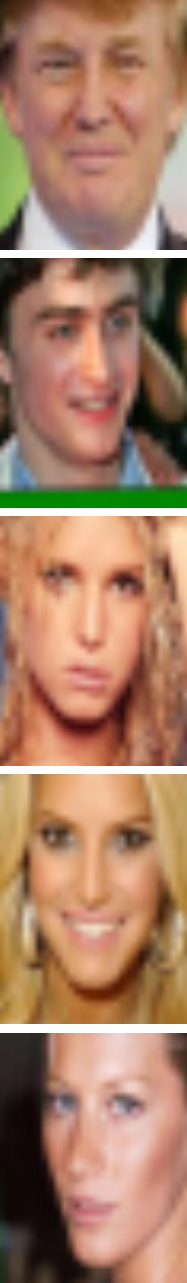}}
\centerline{BICNN}
\end{minipage}
\!
\begin{minipage}{0.1\linewidth}
\centerline{\includegraphics[width=1\textwidth]{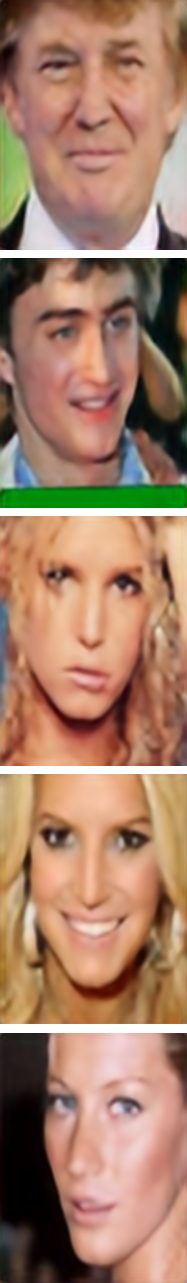}}
\centerline{GLN}
\end{minipage}
\!
\begin{minipage}{0.1\linewidth}
\centerline{\includegraphics[width=1\textwidth]{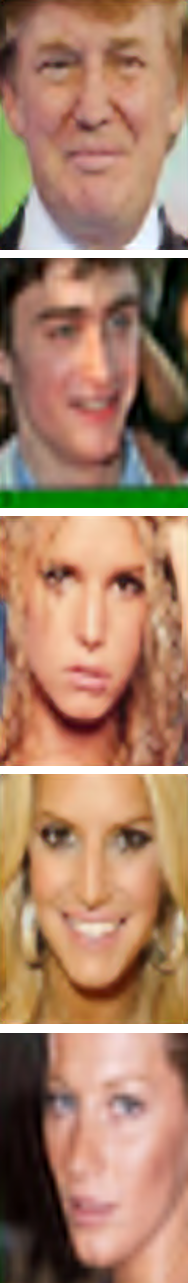}}
\centerline{VDSR}
\end{minipage}
\!
\begin{minipage}{0.1\linewidth}
\centerline{\includegraphics[width=1\textwidth]{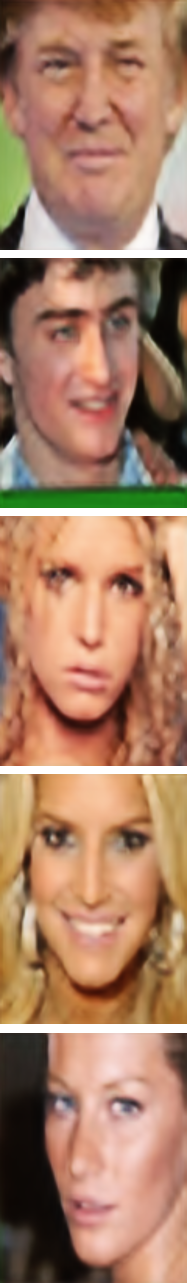}}
\centerline{SRGAN}
\end{minipage}
\!
\begin{minipage}{0.1\linewidth}
\centerline{\includegraphics[width=1\textwidth]{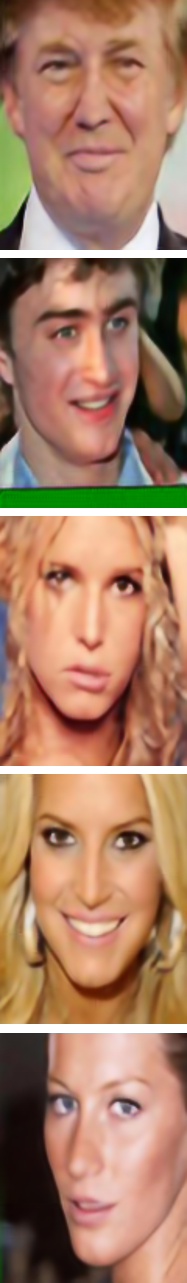}}
\centerline{Our}
\end{minipage}
\!
\begin{minipage}{0.1\linewidth}
\centerline{\includegraphics[width=1\textwidth]{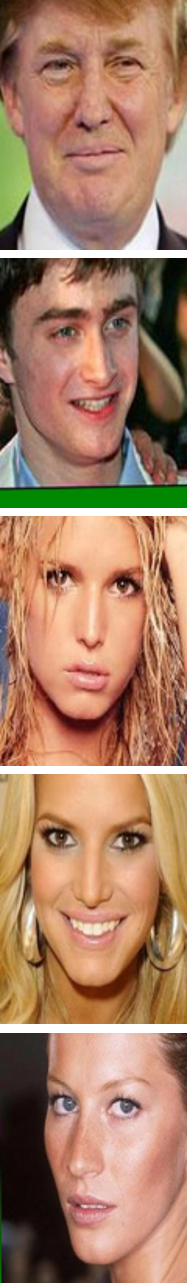}}
\centerline{Original}
\end{minipage}
\caption{Qualitative comparison on the PubFig~\cite{kumar2009attribute} dataset with a scaling factor of 4. The image is best viewed by zooming in on the electronic version.}
\label{fig:pubfig_x4}
\end{figure*}
%%%%%%%%%%%%%%%%%%%%%%%%%%%%%%%%%%%%%%Fig for pubfig x16%%%%%%%%%%%%%%%%%%%%%%%%%%%%%%%%%%%%%%%%%%%%%%%%%%%%%%%%%%%%%%%%%
\begin{figure*}[]
\begin{minipage}{0.1\linewidth}
\centerline{\includegraphics[width=1\textwidth]{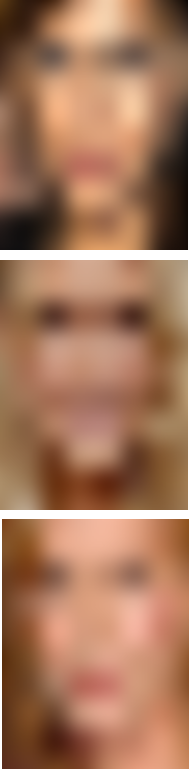}}
\centerline{Bicubic}
\end{minipage}
\!
\begin{minipage}{0.1\linewidth}
\centerline{\includegraphics[width=1\textwidth]{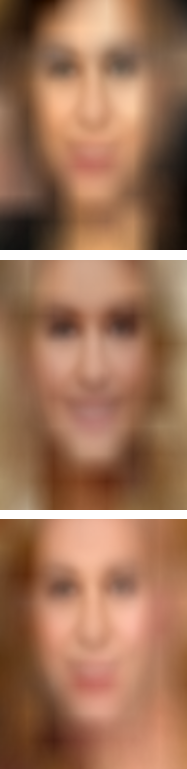}}
\centerline{BICNN}
\end{minipage}
\!
\begin{minipage}{0.1\linewidth}
\centerline{\includegraphics[width=1\textwidth]{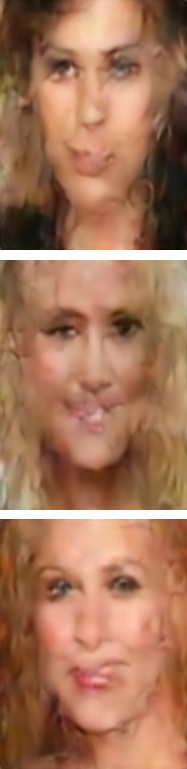}}
\centerline{GLN}
\end{minipage}
\!
\begin{minipage}{0.1\linewidth}
\centerline{\includegraphics[width=1\textwidth]{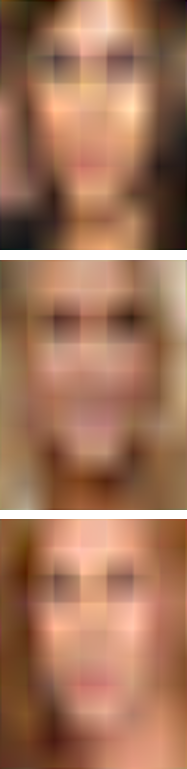}}
\centerline{VDSR}
\end{minipage}
\!
\begin{minipage}{0.1\linewidth}
\centerline{\includegraphics[width=1\textwidth]{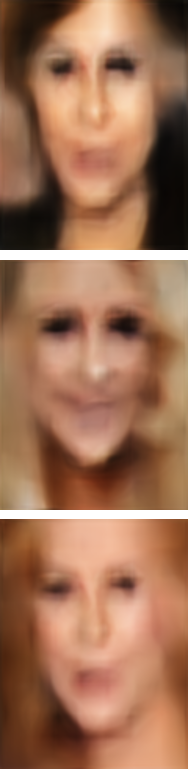}}
\centerline{SRGAN}
\end{minipage}
\!
\begin{minipage}{0.1\linewidth}
\centerline{\includegraphics[width=1\textwidth]{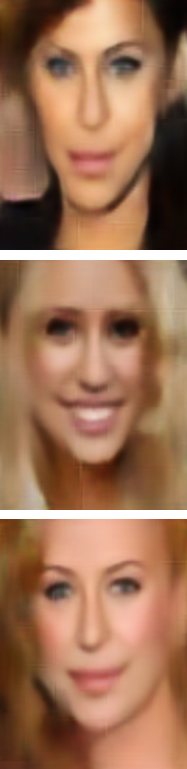}}
\centerline{Our}
\end{minipage}
\!
\begin{minipage}{0.1\linewidth}
\centerline{\includegraphics[width=1\textwidth]{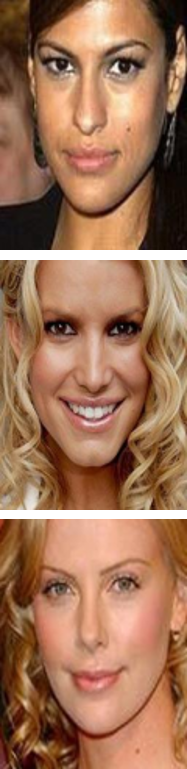}}
\centerline{Original}
\end{minipage}
\caption{Qualitative comparison on the PubFig~\cite{kumar2009attribute} dataset with a scaling factor of 16. The image is best viewed by zooming in on the electronic version.}
\label{fig:pubfig_x16}
\end{figure*}

%\begin{table}[]
%\centering
%\begin{tabular}{|c|c|c|}
%\hline
%Algorithm     & Parameters & PSNR  \\ \hline \hline
%Plain CNN     & 834,033    & 21.94 \\ \hline
%BatchNorm CNN & 835,569    & 23.02 \\ \hline
%EDSR~\cite{edsr}          & 2,463,217  & 23.07 \\ \hline
%Our           & 1,706,850  & 23.62 \\ \hline
%\end{tabular}
%\caption{State-of-the-arts with BatchNorm. We observe that a plain CNN with batch normalization can obtain significant improvement in face hallucination.}
%\label{table:BN_Comparison}
%\end{table}

\subsection{Ablation Study on the Policy Network}
\label{sec:ablation}
\yukai{To demonstrate the effectiveness of the policy network, we compare several variants of our Attention-FH as baseline methods, i.e., ``CNN-16'', ``Our w/o attention'', ``Our w/ random'', ``Our w/ sequences'', ``Our w/o size-free'' and ``Our w/ $I_{0}$ agent''. ``CNN-16'' indicates a plain convolution network with 16 layers. ``Our w/o attention'' refers to that the whole face image is recursively enhanced via a recurrent model from a holistic perspective instead of attending patches. ``Our w/ random'' denotes that we randomly select the attention region to perform random patch-based enhancement. ``Our w/ sequences'' scans through the whole image in sequence. ``Our w/o size-free'' uses a fixed attention box inside the policy network. ``Our w/ $I_{0}$ agent'' replaces $I_t$ with $I_0$ as the input for the policy network. Moreover, we also consider the complicated transformation mechanism that excludes RL, e.g., Attention-FH with spatial transform net (STN)~\cite{st_network} (denoted as ``STN''), which is capable of learning to select informative patches for face hallucination by estimating the subgradients w.r.t the location of the captured patch.
}
%To demonstrate the effectiveness of the policy network, we have compared with several variants of our Attention-FH as baseline methods, i.e., "CNN-16", "Our w/o attention", "Our w/ random", "Our w/o size-free", "Our w/ $I_{0}$ agent". Specifically, CNN-16 indicates a plain convolution network with 16 layers. “Our w/o attention means to recursively enhance the whole face image via a recurrent model from the holistic perspective instead of attending patches. Our w/ random denotes that we randomly select the attention region to form a patch-based random enhancement fashion. Our w/o size-free uses a fixed attention box inside the policy network instead. {\yukai{Our w/ $I_{0}$ agent conducts the low-res face image as inputs for the policy network.}} Moreover, we have also considered the complicated transformation mechanism that excludes reinforcement learning, e.g., Attention-FH with Spatial Transform Net (STN)~\cite{st_network} (denoted as "STN"), which is capable of learning to pick out informative patches for face hallucination by estimating the sub-gradients w.r.t the location of the captured patch.

To demonstrate the superiority of the policy network, we conduct a comparison with the STN~\cite{st_network}. Specifically, our Attention-FH and the STN both employ a paramount patch mining strategy. STN chooses the patch by learning auxiliary terms by minimizing the MSE, and our Attention-FH employs RL to iteratively identify the correct attention region. Therefore, we conduct STN~\cite{st_network} for comparison to illustrate the strengths of RL. For STN, the outputs of the policy network are replaced with a vector $\{x,y\}$, which indicates the corresponding coordinates to extract the patch. We fix the transform scale to ensure an attended patch size of $60 \times 45$. As shown in Tab.~\ref{table:ablation}, our full model outperforms STN by a clear margin. This result confirms that RL is beneficial in crucial region mining for face hallucination.

\begin{table}[]
\footnotesize
\centering
\begin{tabular}{|c|c|c|}
\hline
Algorithm         & LFW $\times$4 & LFW $\times$8 \\ \hline \hline
CNN-16            & 29.11  & 24.02  \\ \hline
Our w/o attention & 32.29  & 25.69  \\ \hline
Our w/ random & 31.63  & 25.74  \\ \hline
Our w/ sequences & 31.78  & 25.98  \\ \hline
Our w/o size-free & 32.89  & 26.12  \\ \hline
Our w/ $I_{0}$ agent           & 32.12  & 25.97   \\ \hline
STN~\cite{st_network} & 28.13  & 25.75  \\ \hline \hline
Our & 33.02 & 26.24  \\ \hline
\end{tabular}
\caption{Comparison of our proposed model under different settings.}
\label{table:ablation}
\end{table}

\subsubsection{Effectiveness of Patch-wise Enhancement} 
As shown in Tab.~\ref{table:ablation}, our full model surpasses ``Our w/o attention'' by 0.73 dB and 0.28 dB on the LFW dataset with the scaling factors of 4 and 8, respectively. This justifies that in-the-wild faces are too changeable to restore, however individual facial parts are relatively stable and can be exploited for partial enhancement. Besides, ``Our w/ random'' obtains a significant improvement over ``CNN-16'', which indicates that the multiple recurrent enhancements itself help to improve the image recovery. Furthermore, our full model achieves consistently higher PSNR values than ``Our w/ random'' due to the crucial patch sequence optimization implemented via RL.

% As shown in Tab.~\ref{table:ablation}, our full model surpasses ``Our w/o attention'' by 0.73 dB and 0.28 dB on the LFW dataset with the scaling factors of 4 and 8, respectively. Therefore, in-the-wild faces are too variable to restore, although individual facial parts are relatively stable. 
% %SENIOR EDITOR: Please ensure that the intended meaning has been maintained in this edit.
% Meanwhile, ``Our w/ random'' achieves a significant improvement over ``CNN-16'' because the template of an in-the-wild face is difficult to represent for conventional fully convolutional neural networks. Moreover, Our full model achieves consistently higher PSNR values than ``Our w/ random'' due to the crucial patch mining implemented via RL. 

%Here we demonstrate the effectiveness of reinforcement learning from 2 aspects; namely, the reinforcement learning algorithm and the sequentially attending patches.

%The spatial transform layer adopt the $\{x,y,s\}$ and the face image produced at previous step as input. Given the $\{x,y,s\}$, the spatial transform layer can extracts the attending patch for enhancement. Meanwhile, the extracted patch is expanded with padding 0s by the spatial transform layer with the transform vector $\{-s*x,-s*y,1/s\}$. With the expanded mean, the extracted patch is feasible to tape back to the low-res face image. We therefore can obtain MSE distance between current restored image and groundtruth image in each step and train the recursive enhancement model in a end-to-end fashion. Apart from the spatial transform layer, the rest parts are as same as our Attention-FH. 

\subsubsection{Effectiveness of Sequentially Attending Patches}
We demonstrate the contribution of sequentially attending patches from the perspective of the attention agent. As illustrated in Tab.~\ref{table:ablation}, we first input the face image without enhancements into the policy network, e.g., ``Our w/ $I_{0}$ agent'', and observe that the performance (PSNR) degrades on the LFW dataset with scaling factors of 4 and 8. By contrast, our full model produces better scores. This result proves that the restored patch not only improves the final face image but also leads the agent to select more accurate region sequence for restoration.

\subsubsection{Effectiveness of Increasing Recursion Depth}
\label{sec:recursio_depth}

% \begin{table}[]
% \centering
% \begin{tabular}{|c|c|c|}
% \hline
% Recurrent Step & LFW x4 & LFW x8 \\ \hline \hline
% T=5            & 29.13  & 24.89  \\ \hline
% T=15           & 32.87  & 26.19  \\ \hline
% T=18           & 32.93  & 26.24  \\ \hline
% T=25           & 32.94  & 26.23  \\ \hline
% T=35           & 32.92  & 26.39  \\ \hline
% \end{tabular}
% \caption{PSNR comparison of the variants of our Attention-FH using different number of steps for sequentially enhancing facial parts on the LFW dataset. It is obvious that $T=15$ achieves a balance between accuracy and efficiency.}
% \label{table:recursion_depth}
% \end{table}

\begin{figure}[t]
    \centering
    \includegraphics[width=0.95\textwidth]{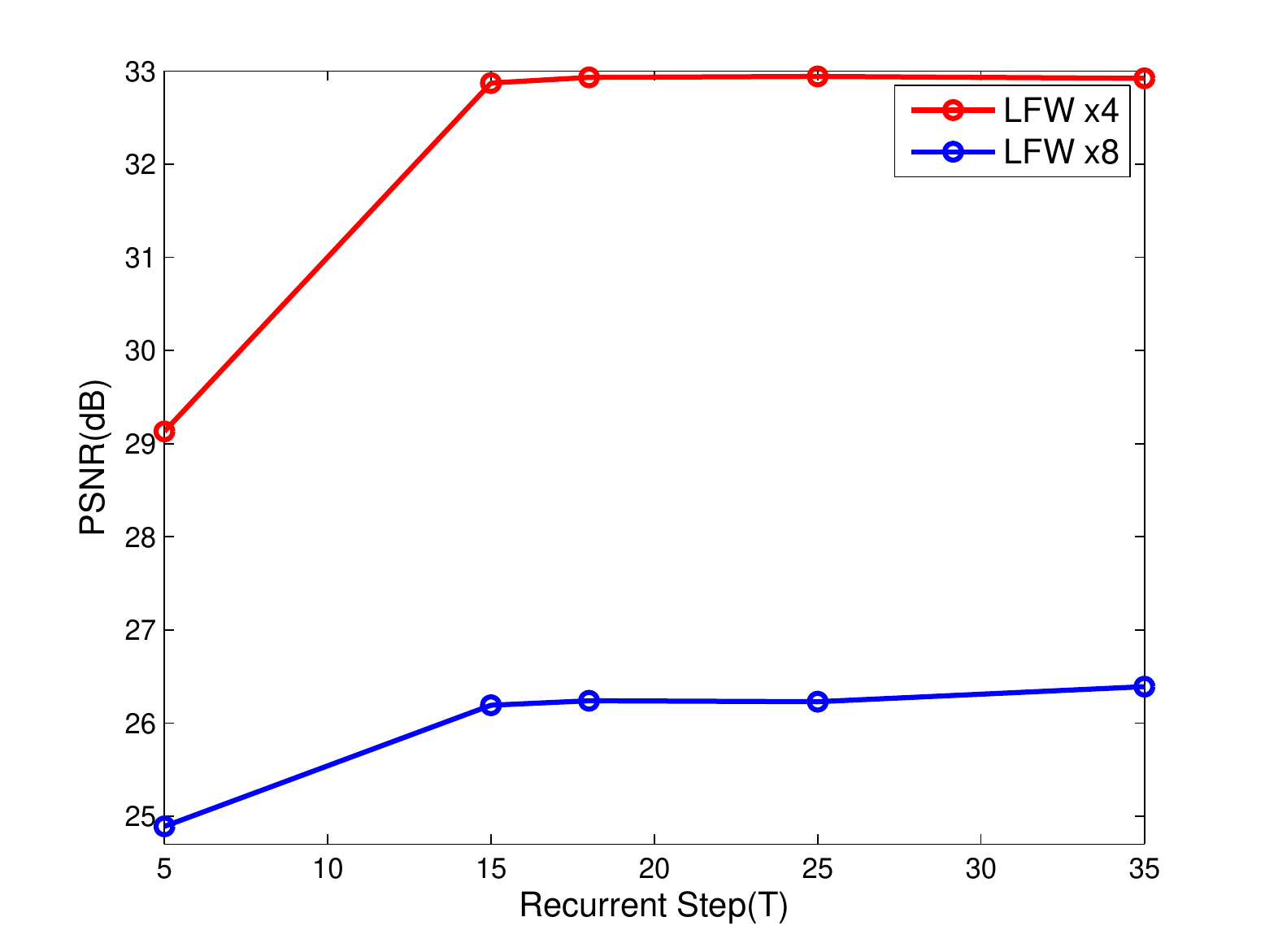}

    \caption{PSNR comparison on the variants of our Attention-FH using different numbers of steps for sequentially enhancing facial parts on the LFW dataset. $T=18$ achieves a balance between accuracy and efficiency.}
    \label{fig:recur_step}
    %\vspace{-3mm}
\end{figure}

To demonstrate the sensitivity of our Attention-FH in terms of the recursive step $T$, we explore the effect of different recursive steps $T$ for sequentially enhancing facial parts. Specifically, we conduct the experiment under five different settings ($T=5,15,18,25,35$) of recursive step on the LFW dataset with scaling factors of 4 and 8. Fig.~\ref{fig:recur_step} shows that the face hallucination performance gradually increases with increasing number of attention steps. The PSNR measure improves dramatically when the number of recursion steps is small, as the extracted patches are still not enough to cover the whole image. When the number of recursion steps reaches more than $15$, the extracted patches are generally enough to cover the whole image. Beyond $15$ steps, the step-wise performance improvement in PSNR becomes negligible. This phenomenon becomes more obvious as the number of steps approaches $25$. Owing to the size-free attention strategy and stabilized reward function, we can obtain considerable restored quality under the PSNR metric when $T$ is set to $18$. In our experiment, we empirically set $T=18$ considering the acceptable computational costs under practical scenarios. 

%\subsubsection{Effectiveness of patch-wise enhancement scheme} 
%We further evaluate the effectiveness of our patch-wise enhancement scheme. In table~\ref{table:Ablation}, ``CNN-16" indicates the results of a 16-layered fully convolution neural network. By comparing our Attention-FH with ``CNN-16", there are 3.82 dB and 2.14 dB improvements in terms of PSNR on LFW of factor 4 and factor 8. We also conduct another ablation study by recurrently enhancing the whole image at each step without extracting patches, named as ``Our w/o attention". This model has the same architecture as our Attention-FH, and the number of recurrent steps is set to 5, which nearly covers the same area of overlapping regions as our full model. From table~\ref{table:Ablation}, we can see that although the recurrent model without attention can achieve promising results, our Attention-FH with attention can still promote 0.67 dB and 0.46 dB on LFW $4\times$ and LFW $8\times$ respectively, which demonstrates the effectiveness of using attention-aware model and reinforcement learning.

\begin{figure}[t]
    \centering
    \includegraphics[width=0.95\textwidth]{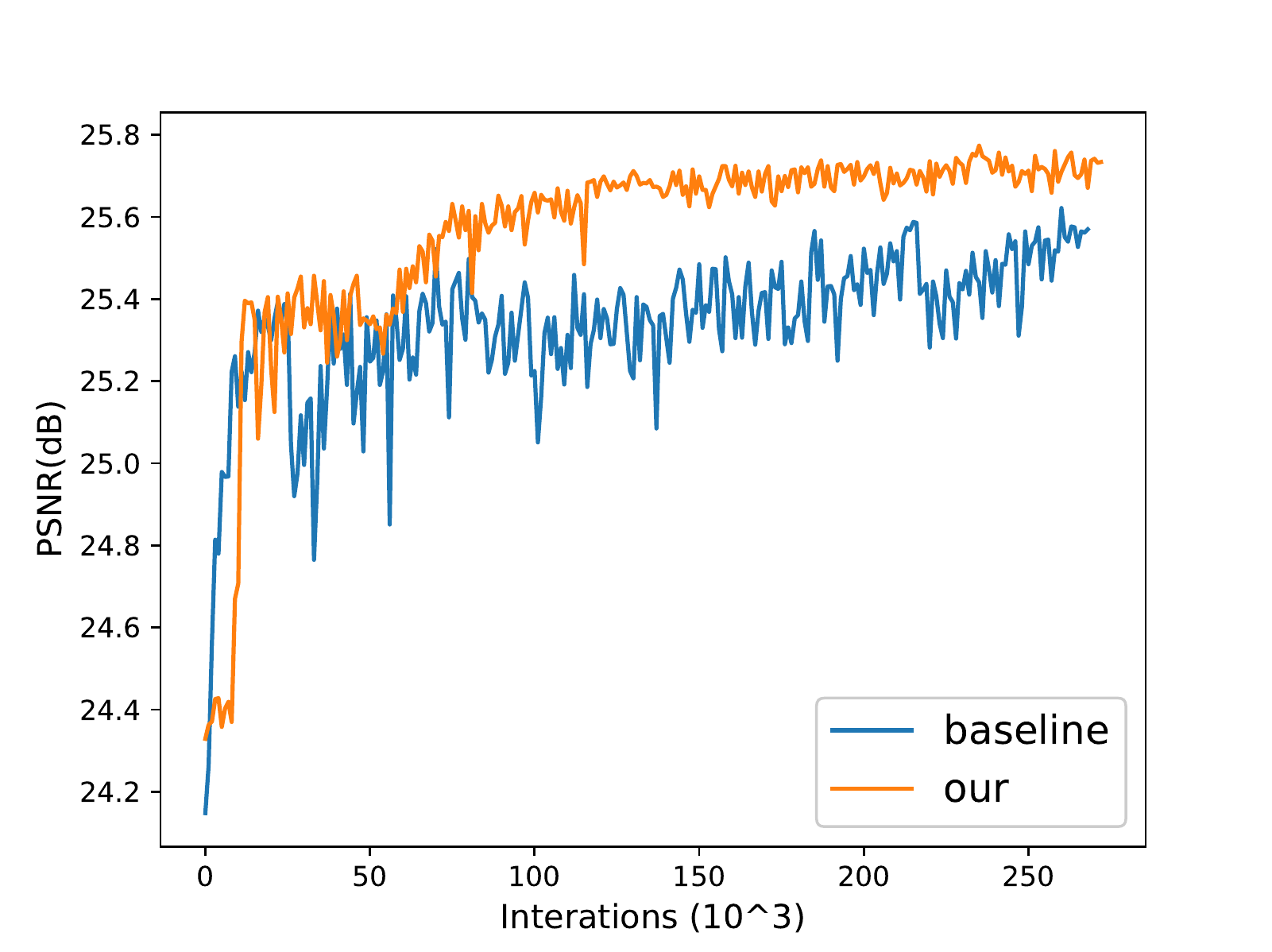}

    \caption{We enhance the reward function by improving its stability. We redefine the previous reward function from our preliminary conference version by means of a more accurate baseline and achieve superior performance.}\label{fig:stability}
    %\vspace{-3mm}
\end{figure}

\subsubsection{Effectiveness of the Stabilized Reward Function}
We conduct an experiment to validate the contribution of the proposed reward function. Compared with former reward function, our stabilized reward function incorporates PSNR gain value instead of absolute PSNR value as the reward. Given the renewal reward, Attention-FH demonstrates an accurate attention agent towards face hallucination. As shown in Fig.~\ref{fig:stability}, our reward function achieves more stable performance with lower variance. Furthermore, compared with the reward function from our preliminary conference version, this new reward function leads to higher and more stable PSNR results for our Attention-FH.

%We conduct an experiment to validate the contribution of the proposed reward function. As shown in Fig.~\ref{fig:stability}, our reward function achieves stable performance with low variance. Furthermore, compared with the reward function from our preliminary conference version, this new reward function leads to higher and more stable PSNR results for our Attention-FH.

\subsubsection{Effectiveness of the Size-free Attention Mechanism}
As depicted in Fig.~\ref{fig:ratio}, our facial component has a different size for each identity.
%SENIOR EDITOR: Please ensure that the intended meaning has been maintained in this edit.
We improve our policy network by implementing a flexible attention mechanism to attend the facial part accurately. We conduct an ablation study by adopting a fixed attention box in the policy network, named ``Our w/o size-free'', to validate the effectiveness. This model has the same settings as our full model, except for using a 60 $\times$ 60 attention box. Tab.~\ref{table:ablation} shows that although ``Our w/o size-free'' produces favorable results, our Attention-FH achieves 0.13 dB and 0.12 dB improvements with scaling factors of 4 and 8, respectively. These results confirm the contribution of the proposed size-free attention mechanism.

\subsection{Ablation Study on the Local Enhancement Network}
\label{sec:local_enhancement_comparison}

\begin{table}[]
\centering
\footnotesize
\begin{tabular}{|c|c|c|}
\hline
Algorithm & PSNR  & Time (millisecond) \\ \hline \hline
VDSR~\cite{vdsr}      & 23.56 & 0.312   \\ \hline
Sub-pixel~\cite{subpixel} & 23.59 & 0.066  \\ \hline
GLN~\cite{gln}       & 23.63 & 0.092  \\ \hline
LapSRN~\cite{laplacian} & 23.64 & 0.081  \\ \hline
U-Net~\cite{unet}     & 23.60 & 0.057  \\ \hline
FSRCNN~\cite{fsrcnn}    & 23.64 & 0.075  \\ \hline
\end{tabular}
\caption{Experimental study of the trade-off between efficiency and accuracy on different local enhancement network architectures. This analysis is conducted on images of size 120$\times$160.}
\label{table:local_enhancement_comparison}
\end{table}

Since the pipeline of our Attention-FH is flexible and extensional, we investigate the use of different network architectures as the local enhancement network. To make the investigation comprehensive, we consider several methods, namely, VDSR~\cite{vdsr}, Sub-pixel~\cite{subpixel}, GLN~\cite{gln}, LapSRN~\cite{laplacian}, U-Net~\cite{unet} and FSRCNN~\cite{fsrcnn}. For U-Net~\cite{unet}, we reduce the parameters to avoid the case that the model is too large to train. 
%For VDSR~\cite{vdsr}, we add it with batch normalization after each convolution layer. 
As shown in Tab.~\ref{table:local_enhancement_comparison}, LapSRN~\cite{laplacian} achieves the best results while U-Net~\cite{unet} is the most efficient method for generating hallucinated faces. However, neither method exhibits distinct differences in terms of PSNR under the recurrent attention mechanism. We choose FSRCNN~\cite{fsrcnn} as the implementation of our local enhancement network based on the trade-off between efficiency and accuracy.

\begin{table}[t]
\footnotesize
\centering
\begin{tabular}{|c|c|c|c|}
%\hline
\hline
Method         & Parameter  & PSNR & Time \\ 
& Number & & (frame/second) \\
\hline 
\hline
VDSR~\cite{vdsr}  &  664,704  & 21.94  & 0.025 \\ 
EDSR~\cite{edsr}  & 2,463,217  & 24.21 & 0.045 \\ 
\hline
Our           & 1,706,850  & 23.62 & 0.092 \\ 
\hline
\end{tabular}
\caption{Efficiency comparison on the PubFig dataset with a scaling factor of 8.}
\label{table:efficiency}
\end{table}

\begin{figure*}[htb] % htbp
	\centerline{\includegraphics[width=0.8\textwidth]{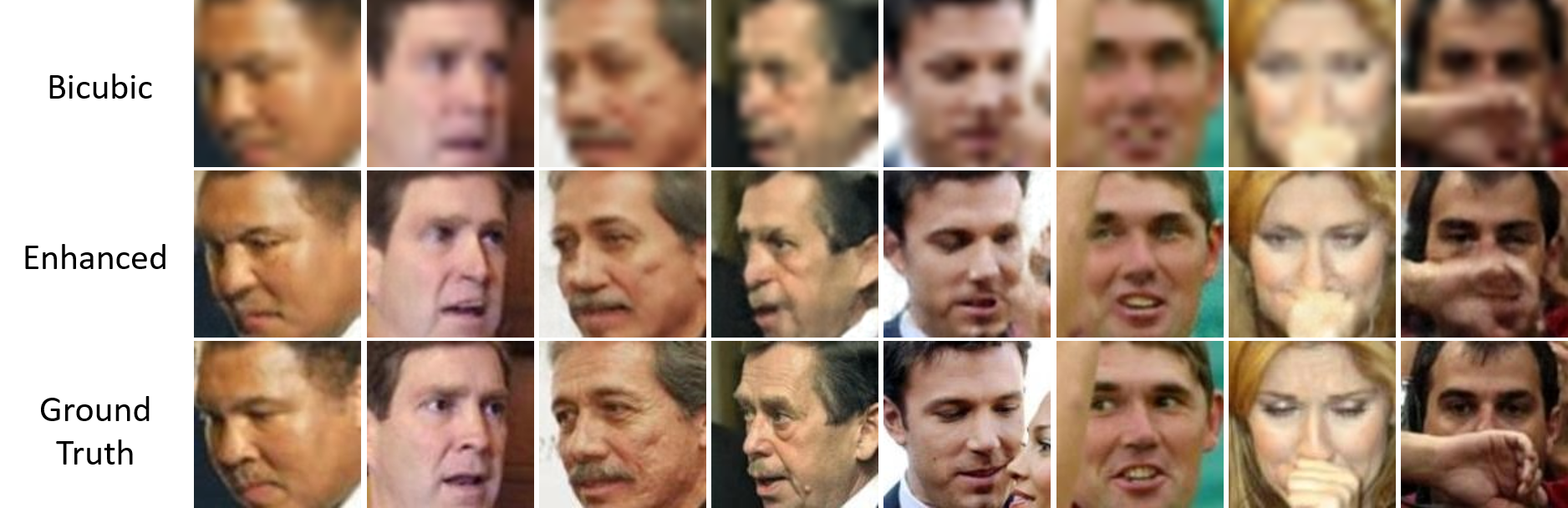}}
	\caption{Visual results on profile$\backslash$occlusion samples. We conduct this comparison on LFW~\cite{lfw-funneled} with $\times$8 factor. Best viewed by zooming in the electronic version.}
	\label{fig:limitation}
%\vspace{-3mm}
\end{figure*}
\subsection{Efficiency Analysis}
\syk{We have also conducted an experimental comparison to verify the efficiency of Attention-FH. The results in Tab.~\ref{table:efficiency} demonstrate that our Attention-FH requires very little time cost to achieve the superior performance. Compared with EDSR, our model achieves remarkable parameter advantages. As Attention-FH is less efficient than VDSR, the proposed model achieves significant improvement over restoration quality. With a single GPU, Attention-FH is able to perform real-time efficiency. However, Attention-FH still meets an efficiency bottleneck when it is deployed on the mobile platform.
}
\subsection{Limitations}
\syk{In this section, we discuss the limitation of Attention-FH. With the novel attention mechanism, our model is capable of hallucinating profile$\backslash$occlusion faces well. However, Attention-FH may hallucinate the incorrect facial part if the occluded content is close to facial component. As shown in the last row of Fig.~\ref{fig:limitation}, Attention-FH hallucinate a mouth in the occlusion area, which occurred by hand. This failure case illustrates the limitation that Attention-FH meets an upper bound on complex occlusion samples. We will further improve Attention-FH by enhancing the generalization ability towards complex occlusion cases.
}
\section{Conclusion} \label{sec:conclude}
In this paper, we have proposed a deep RL-based attention mechanism to address the problem of face hallucination. In contrast to traditional patch-wise face hallucination models that usually neglect the interdependency between facial components, our framework implements a deep RL model and jointly optimizes a recurrent policy network, which learns to determine an ordered patch hallucination sequence, and a local enhancement network for facial part super-resolution. Our Attention-FH fully reflects the human visual perception mechanism and is capable of adaptively inferring an optimal search path for each facial image according to its unique appearance features. Extensive experiments show that Attention-FH outperforms state-of-the-art face hallucination methods and achieves leading performance on both widely used evaluation protocols and visual quality comparisons.

\bibliographystyle{IEEEtran}
\bibliography{egbib}

\vspace{-10mm}
\begin{IEEEbiography}[{\includegraphics[width=1in,height=1.25in,clip,keepaspectratio]{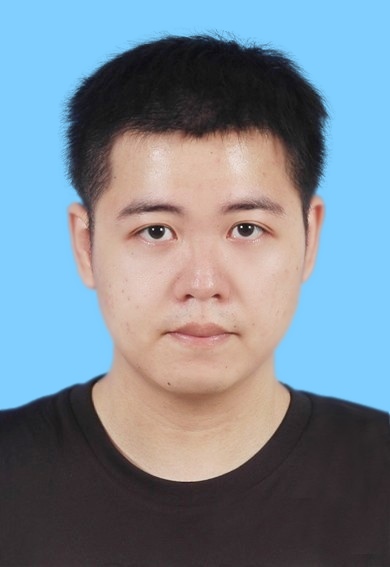}}]{Yukai Shi} received his B.E. degree from Heilongjiang University, Harbin, China, in 2014, and is currently working towards a Ph.D. degree at the School of Data and Computer Science, Sun Yat-Sen University, Guangzhou, China. His research interests includes computer vision and machine learning.
\end{IEEEbiography}

\vspace{-10mm}
\begin{IEEEbiography}[{\includegraphics[width=1in,height=1.25in,clip,keepaspectratio]{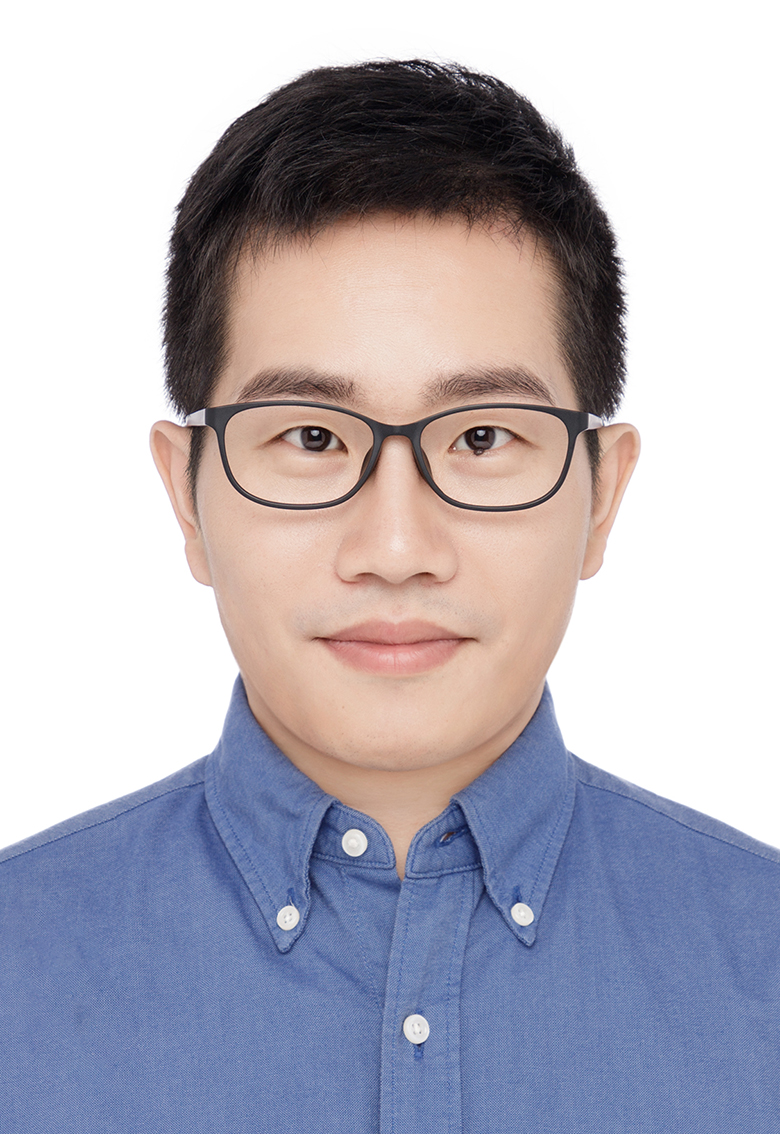}}]{Guanbin Li} (M'15) is currently an associate professor in School of Data and Computer Science, Sun Yat-sen University. He received his PhD degree from the University of Hong Kong in 2016. He was a recipient of Hong Kong PhD Fellowship. His current research interests include computer vision, image processing, and deep learning. He has authorized and co-authorized on more than 30 papers in top-tier academic journals and conferences. He serves as an area chair for the conference of VISAPP. He has been serving as a reviewer for numerous academic journals and conferences such as TPAMI, TIP, TMM, TC, CVPR, AAAI and IJCAI.
\end{IEEEbiography}

\vspace{-10mm}
\begin{IEEEbiography}[{\includegraphics[width=1in,height=1.25in,clip,keepaspectratio]{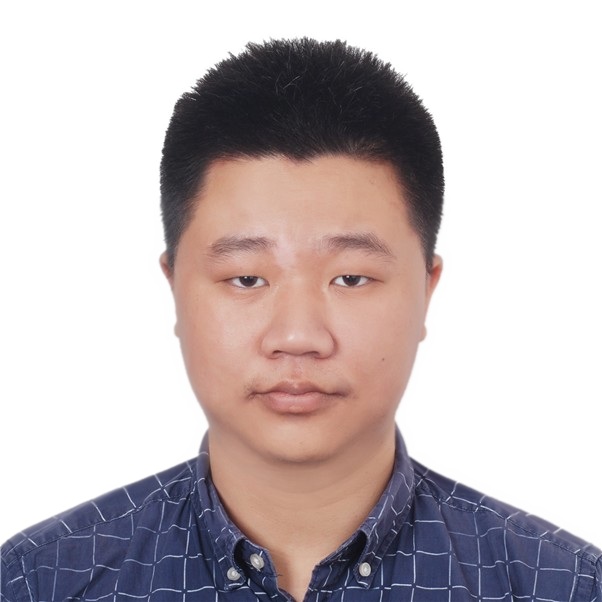}}]{Qingxing Cao} received his B.S. degree in software engineering from Sun Yat-Sen University, Guangzhou, China, in 2013. He is currently pursuing a Ph.D. degree in computer science and technology at Sun Yat-Sen University, advised by Professor Liang Lin.  His current research interests include deep learning and computer vision.
\end{IEEEbiography}

\vspace{-10mm}
\begin{IEEEbiography}[{\includegraphics[width=1in,height=1.25in,clip,keepaspectratio]{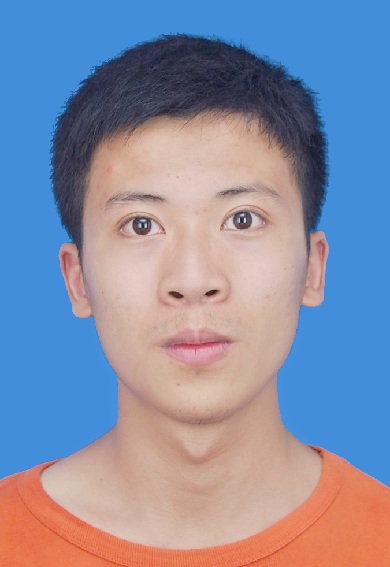}}]{Keze Wang} received his B.S. degree in software engineering from Sun Yat-Sen University, Guangzhou, China, in 2012. He is currently pursuing a dual Ph.D. degree at Sun Yat-Sen University and Hong Kong Polytechnic University, advised by Prof. Liang Lin and Lei Zhang. His current research interests include computer vision and machine learning. More information can be found on his personal website \url{http://kezewang.com}.
\end{IEEEbiography}

\vspace{-10mm}
\begin{IEEEbiography}[{\includegraphics[width=1in,height=1.25in,clip,keepaspectratio]{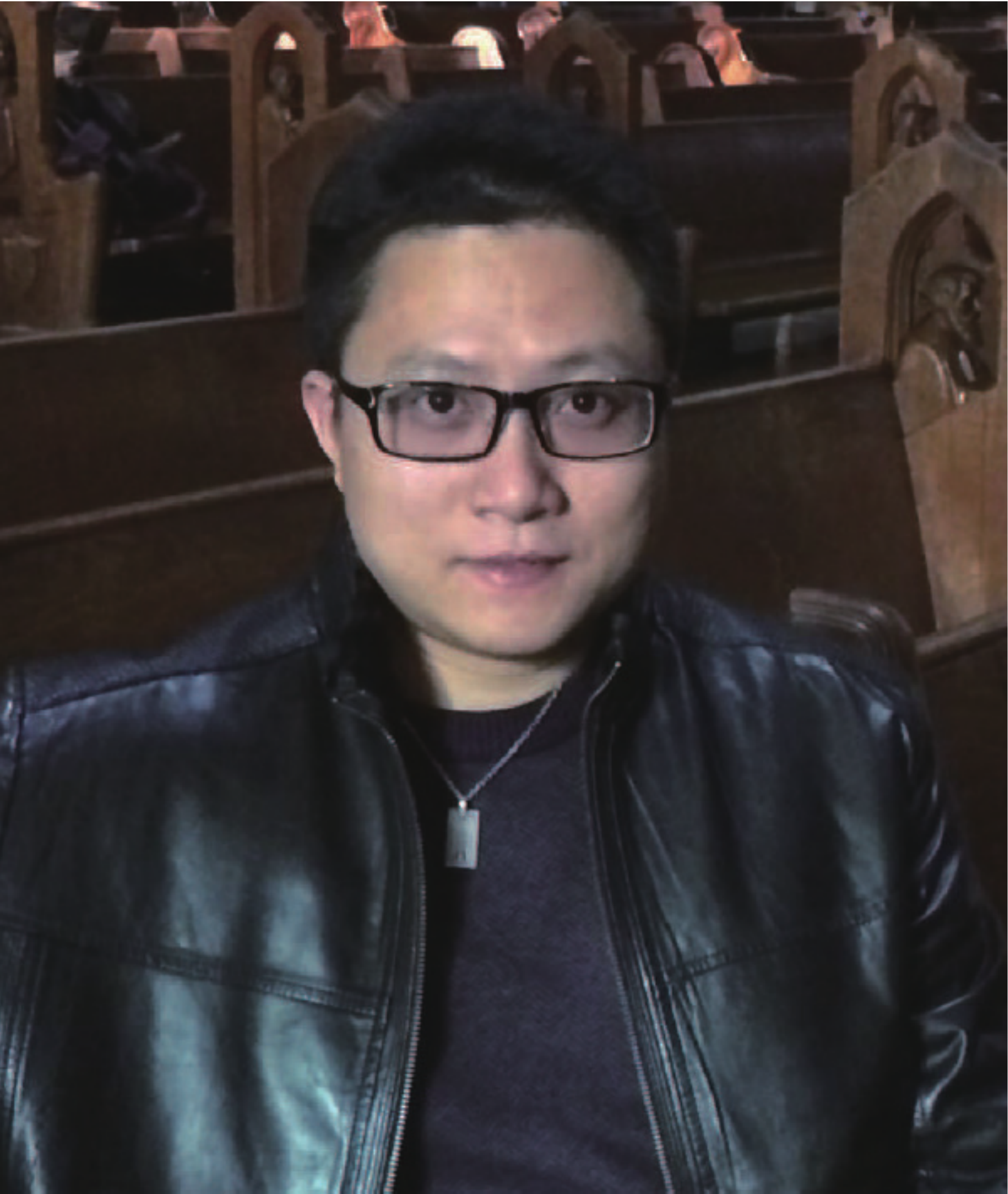}}]{Liang Lin}
(M'09, SM'15) is a full Professor of Sun Yat-sen University. He is the Excellent Young Scientist of the National Natural Science Foundation of China. From 2008 to 2010, he was a Post-Doctoral Fellow at the University of California, Los Angeles. From 2014 to 2015, as a senior visiting scholar, he was with the Hong Kong Polytechnic University and the Chinese University of Hong Kong. He currently leads the SenseTime R$\&$D teams to develop cutting-edge and deliverable solutions on computer vision, data analysis and mining, and intelligent robotic systems. He has authored and co-authored more than 100 papers in top-tier academic journals and conferences. He has been serving as an associate editor of IEEE Trans. Human-Machine Systems, The Visual Computer and Neurocomputing. He served as Area/Session Chair for numerous conferences, including ICME, ACCV, and ICMR. He was the recipient of the Best Paper Runners-Up Award at ACM NPAR 2010, a Google Faculty Award in 2012, the Best Paper Diamond Award at IEEE ICME 2017, and the Hong Kong Scholars Award in 2014. He is a Fellow of IET.
\end{IEEEbiography}

\end{document}